\pdfoutput=1

\documentclass[11pt]{article}

\usepackage[]{acl}

\usepackage{times}
\usepackage{latexsym}

\usepackage[T1]{fontenc}

\usepackage[utf8]{inputenc}

\usepackage{microtype}

\usepackage{inconsolata}

\usepackage{array}
\usepackage{tabularx}
\usepackage{graphicx}
\usepackage{longtable}

\newcommand{\conceptname}{\textsc{Mixed-Session Conversation}}
\newcommand{\dataname}{\textsc{MiSC}}
\newcommand{\memoryname}{Egocentric Memory}
\newcommand{\modelname}{\textsc{EMMA}}

\title{Mixed-Session Conversation with Egocentric Memory}

\author{Jihyoung Jang\quad Taeyoung Kim\quad Hyounghun Kim \\
  Artificial Intelligence Graduate School, UNIST \\
  \texttt{\{jihyoung, taeyoung.kim, h.kim\}@unist.ac.kr} \\}

\begin{document}
\maketitle

\begin{abstract}

Recently introduced dialogue systems have demonstrated high usability. However, they still fall short of reflecting real-world conversation scenarios. Current dialogue systems exhibit an inability to replicate the dynamic, continuous, long-term interactions involving multiple partners. This shortfall arises because there have been limited efforts to account for both aspects of real-world dialogues: deeply layered interactions over the long-term dialogue and widely expanded conversation networks involving multiple participants. As the effort to incorporate these aspects combined, we introduce \conceptname{}, a dialogue system designed to construct conversations with various partners in a multi-session dialogue setup. We propose a new dataset called \dataname{} to implement this system. The dialogue episodes of \dataname{} consist of 6 consecutive sessions, with four speakers (one main speaker and three partners) appearing in each episode. Also, we propose a new dialogue model with a novel memory management mechanism, called \textbf{E}gocentric \textbf{M}emory Enhanced \textbf{M}ixed-Session Conversation \textbf{A}gent (\modelname{}). \modelname{} collects and retains memories from the main speaker's perspective during conversations with partners, enabling seamless continuity in subsequent interactions. Extensive human evaluations validate that the dialogues in \dataname{} demonstrate a seamless conversational flow, even when conversation partners change in each session. \modelname{} trained with \dataname{} is also evaluated to maintain high memorability without contradiction throughout the entire conversation.\footnote{Our dataset/code are publicly available at \url{https://mixed-session.github.io/}}

\end{abstract}

\section{Introduction}
\begin{figure*}[t]
    \centering
    \includegraphics[width=1.95\columnwidth]{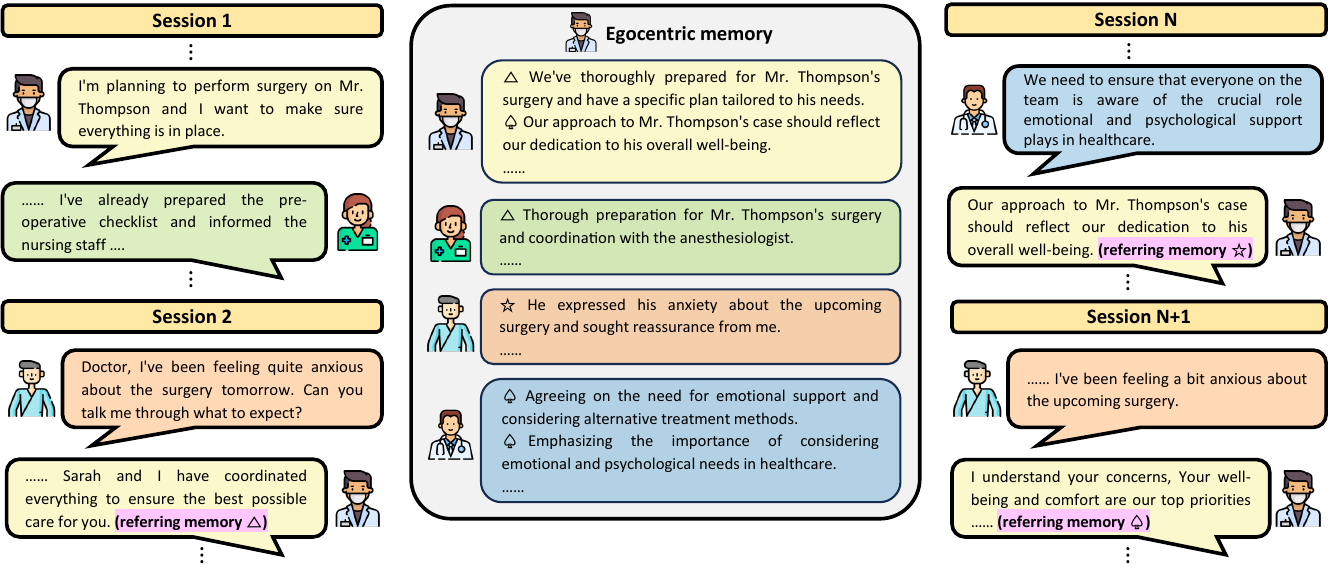}
    \caption{A sample of our \dataname{}. The main speaker collects each speaker's memory from the main speaker's perspective at the end of each session and utilizes this memory to proceed with the conversation in the following session. The memory referenced when generating utterances can be identified through symbols, and connected memories are represented by the same symbol.}
    \label{fig:MainExam}
    \vspace{-5pt}
\end{figure*}

Dialogue systems have been evolving along two dimensions: depth, for supporting long-term interactions, and width, for accommodating a greater number of conversation partners. Multi-session conversations~\cite{xu-etal-2022-beyond, bae-etal-2022-keep, jang-etal-2023-conversation, zhang-etal-2023-mind} have been proposed as an instance of such long-term dialogue systems retaining dialogue context across consecutive sessions.
Expanding the network of conversation partners, in the other dimension, includes multi-party conversations~\cite{ouchi-tsuboi-2016-addressee, poria-etal-2019-meld, le-etal-2019-speaking, wang-etal-2020-response, mahajan-shaikh-2021-need, gu-etal-2021-mpc, wei2023multiparty, chen-etal-2023-places, gu-etal-2023-madnet}. It expands the scope of interactions by increasing the number of conversation partners engaged in a dialogue session.

However, in the real-world, conversations occur within complex contexts that are both lengthy and deeply layered, involving a wide range of people. Therefore, focusing on either of the two dimensions would not fully capture these dynamics. Given this significance, there have been surprisingly few efforts to advance dialogue systems in both directions.

To expand the boundaries of those dialogue systems, we introduce \conceptname{}.
Unlike multi-session conversations, where a speaker engages with one fixed partner across all sessions, the main speaker in \conceptname{} encounters multiple partners in a mixed order of sessions. This approach is thus referred to as Mixed-Session.
Specifically, \conceptname{} consists of multiple dialogue sessions, during which several speakers, including a main speaker, interact dynamically over time. The main speaker engages in conversations with different partners, one partner per session, focusing on a specific event. This setting enables a dialogue system to build a deep, layered context with each of its partners, thereby expanding and complicating the dynamics.

To implement \conceptname{}, we develop a dialogue dataset named \dataname{} (Figure~\ref{fig:MainExam}). \dataname{} comprises 8.5K episodes, with each episode consisting of 6 sessions (a total of 51K sessions). In each episode, four speakers participate, with one main speaker involved in all sessions and each of the other three speakers participating as a conversation partner. To enable the main speaker to retain all contexts across sessions and partners, we introduce a new memory managing system called \memoryname{}. \memoryname{} keeps memory about each partner from the main speaker's perspective, enabling accurate recall to align the events with each partner without contradiction. 

We actualize \conceptname{} through a novel dialogue model, named \textbf{E}gocentric \textbf{M}emory Enhanced \textbf{M}ixed-session Conversation \textbf{A}gent (\modelname{}). Trained on \dataname{}, \modelname{} ensures seamless continuity during interactions between speakers leveraging \memoryname{}. 
As the session progresses, the memory of each speaker is newly added or updated; thereby, all memory can be retained without losing information about the previous sessions and partners.

Through extensive human evaluation, the quality of \dataname{} and conversations generated from \modelname{} are verified to have high qualities. To be specific, \dataname{} exhibits high consistency and coherence throughout the episode, retaining accurate memory of each partner from the main speaker's perspective, even with conversation partners changing with each session. Conversations from \modelname{} demonstrate high humanness, engagingness, and memorability.

Our contributions in this study are:
\begin{enumerate}
\itemsep0em 

\item We introduce \dataname{}, which consists of 6 dialogue sessions and four speakers per episode, implementing \conceptname{}.

\item We propose \modelname{}, a novel dialogue model enabling seamless continuity for subsequent sessions based on \memoryname{}.

\item In extensive human evaluations, our \dataname{} and \modelname{} demonstrate high consistency and coherence, ensuring natural continuity with the conversation partner changing at each session.

\end{enumerate}

\section{Related Works}

\begin{figure*}[t]
    \centering
    \includegraphics[width=1.98\columnwidth]{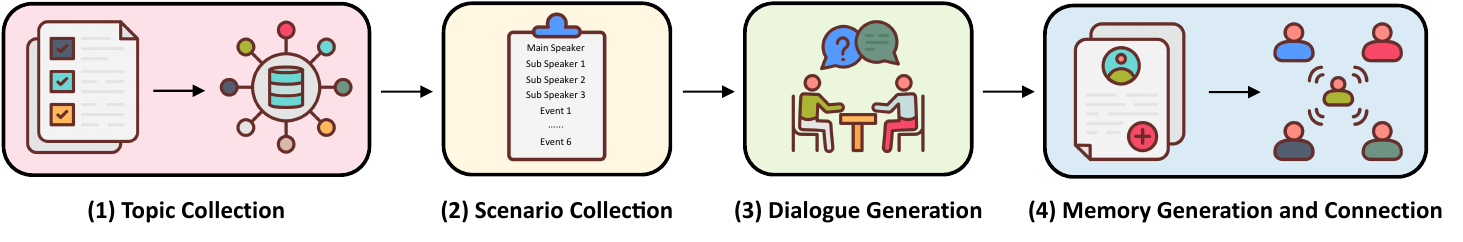}
    \caption{Overall pipeline for constructing the \dataname{}.}
    \label{fig:DataMet}
    \vspace{-10pt}
\end{figure*}

\paragraph{Multi-Session Conversations.}
One direction in which dialogue systems have been pushing is enabling long-term interaction. Multi-session conversation~\cite{xu-etal-2022-beyond} is one of the systems enabling such long-term conversation. Moreover, there are attempts to effectively manage memory~\cite{bae-etal-2022-keep} and implement longer time intervals and relationships between speakers in multi-session conversations~\cite{jang-etal-2023-conversation}. However, to our knowledge, there is no existing research that explores changing conversation partners with each session. Our \conceptname{} is the first system to involve multiple partners in a multi-session setting.

\paragraph{Multi-Party Conversations.}
Many research efforts on dialogue systems have been made to expand the range of conversational partners. While previous dialogue systems have mainly focused on building conversational systems between two speakers, recent research has shifted its focus towards multi-party conversation setup, which are more prevalent in real-life dialogues~\cite{ouchi-tsuboi-2016-addressee, poria-etal-2019-meld, le-etal-2019-speaking, wang-etal-2020-response, mahajan-shaikh-2021-need, gu-etal-2021-mpc, wei2023multiparty, chen-etal-2023-places, gu-etal-2023-madnet}. However, there have yet to be dialogue systems that simultaneously accommodate both multi-session and multi-party aspects. In this study, we propose \dataname{} and \modelname{}, which we hope will serve as pioneering contributions to the open-domain dialogue field as a dataset and model, respectively.

\paragraph{Machine Generated Datasets.}
In previous research, dataset generation largely depended on crowdsourcing, involving human participants to manually produce data according to specific guidelines. This method is labor-intensive, costly, and might result in inconsistent data quality due to varying performance among crowd-workers. To address these issues, recent studies have explored using machines, specifically large language models (LLMs) like GPT, for data generation~\cite{kim-etal-2022-prosocialdialog, zheng-etal-2023-augesc, kim-etal-2023-soda, jang-etal-2023-conversation, xu-etal-2023-baize}. This approach is seen as more efficient and cost-effective, allowing for high-quality data production with precise control over the process through well-designed prompts, ensuring consistency. Notably, it has been found that even data for complex scenarios, which are difficult for humans to handle, can be generated while maintaining high quality~\cite{Gilardi_2023}.
\section{\dataname{}{}}

We introduce \conceptname{}, a novel dialogue setting that advances along both depth and width dimensions simultaneously. Unlike previous systems, \conceptname{} allows the main speaker to engage with different sub-speakers as conversation partners in each session. To implement this conversation system, we propose a new dataset called \dataname{}.

\dataname{} comprises 8.5K episodes, each consisting of 6 sessions, totaling 51K sessions in all. In each episode, four speakers are involved, with one acting as the main speaker and the others as sub-speakers.

We construct the dialogue dataset in a sequential manner, starting with the collection of topics and progressing to the generation of conversations. We use LLMs to build the dataset through elaborately designed methodologies (please refer to Figure~\ref{fig:DataMet} for an overview of the process).

\subsection{Scenario Setup}
To build our dataset, we establish conversational scenarios for each episode. Each scenario includes information about the speakers (names, jobs, or relationships) and a specific event for each session, thus, a total of six events. Our preliminary research shows that the quality of scenarios has a significant impact on the overall dialogue quality. When high-quality scenarios are provided, the difference in dialogue quality between GPT-4 and GPT-3.5 becomes minimal. Accordingly, we utilize GPT-4 to generate the scenarios and GPT-3.5 for the subsequent processes. We generate the episode scenario as follows.

\paragraph{Topic Collection.} 
We generate topics from keywords related to daily life (e.g., health, travel, education, etc.), using them as seeds to generate scenarios. We instruct GPT-4~\cite{achiam2023gpt} to generate topics from a single keyword. For example, topics could be generated for the `food' keyword, such as ``Dishes from My Grandmother's Kitchen'', ``Feasting with Friends: Tales from a Supper Club'', ``Recipe for Love'', and so on.

\paragraph{Scenario Collection.} 
We gather scenarios based on pre-defined topics and generate details about speakers and events related to each topic. Specifically, we ask GPT-4 to create the names, jobs, or relationships of the main speaker and three other participants, as well as seamlessly connected events for each conversation session.

Additionally, we request clear identification of the conversational partner involved in each session event to track the change of partners. The generated scenario serves as a foundation for building each episode. Please see Appendix~\ref{sec:SceExam} for complete scenario examples.

\subsection{Dialogue Generation}
We generate conversations sequentially with ChatGPT~\cite{chatgpt} from the first to the sixth session, each featuring its own unique session event. Given this setup, it is crucial to ensure continuity between sessions by reflecting on the history of the previous sessions in the subsequent one. To accomplish this goal, we employ two methods: session summaries and the main speaker's memory.

We follow \citet{jang-etal-2023-conversation}'s approach to generate summaries. The main speaker's memory is used to retain the content shared with each partner from the main speaker's perspective, as detailed in Section~\ref{sec:egomemory}. Through the integration of these two approaches, we ensure seamless transitions between sessions and facilitate a more cohesive exchange of ideas. Please refer to Appendix~\ref{sec:DataExam} for complete episode examples.

\subsection{\memoryname{}}
\label{sec:egomemory}
We utilize the main speaker's memory to uphold the history of previous sessions. This involves summarizing and preserving memories about each partner and the main speaker themselves, from the main speaker's viewpoint. We refer to this memory management approach as \memoryname{}. It is distinct from previous summarization or memory mechanisms, in that it effectively stores memories related to each partner and establishes links between updated memories across multiple sessions.

\paragraph{Memory Generation.}
At the end of each session, we ask ChatGPT to identify significant events, experiences, appointments, and the emotions expressed during the conversation as memory elements from the main speaker's perspective. These memories are generated and recorded separately for both the main speaker and their partner, incorporating references to previous sessions to ensure continuity and coherence.

\paragraph{Memory Connection.}
\memoryname{} is maintained across multiple sessions to help the main speaker to understand the conversation comprehensively. However, if these memories are disconnected and independent, it may fail to integrate contextual references of similar memories or update changed situations. To effectively manage these memory instances, we ask ChatGPT to connect them. Initially, we connect memory instances within the session and link them with memories from previous sessions. As far as we know, the ability to connect memories and continuously update them is unique to \dataname{}. These interconnected memories are structured into a wide-layered network, providing expanded context to the main speakers.

\paragraph{Memory Tagging.}
To maintain coherence and continuity in each conversation session, memory referencing is employed during the dialogue generation step. This process ensures that each utterance is generated with reference to relevant memories, thereby enhancing the natural flow of the dialogue. Consequently, we assign a corresponding memory reference tag to each utterance. In each session, all utterances from the main speaker, together with the list of memories, are provided as input to ChatGPT. Based on this input, ChatGPT associates each utterance with the corresponding memory index it references. For detailed examples of memory usage in \dataname{}, please see Appendix~\ref{sec:DataExam}.

\begin{table}[t]
\centering
\resizebox{0.9\linewidth}{!}{
\begin{tabular}{l >{\centering\arraybackslash}m{0.38\linewidth}}
\hline
\textbf{Type} & \textbf{Count}\\
\hline
\# of Episodes & 8,556\\
\# of Sessions & 51,336\\
\# of Unique Speaker Name & 4,984\\
\# of Unique Speaker Job & 25,571\\
Avg. Turns per Episode & 46.97\\
Avg. Memory per Episode & 21.26\\
Avg. Memory Links per Episode & 9.49\\
\hline
\end{tabular}
}
\caption{Statistics of \dataname{}.}
\label{tab:DataStat}
\end{table}

\begin{figure*}[t]
    \centering
    \includegraphics[width=1.98\columnwidth]{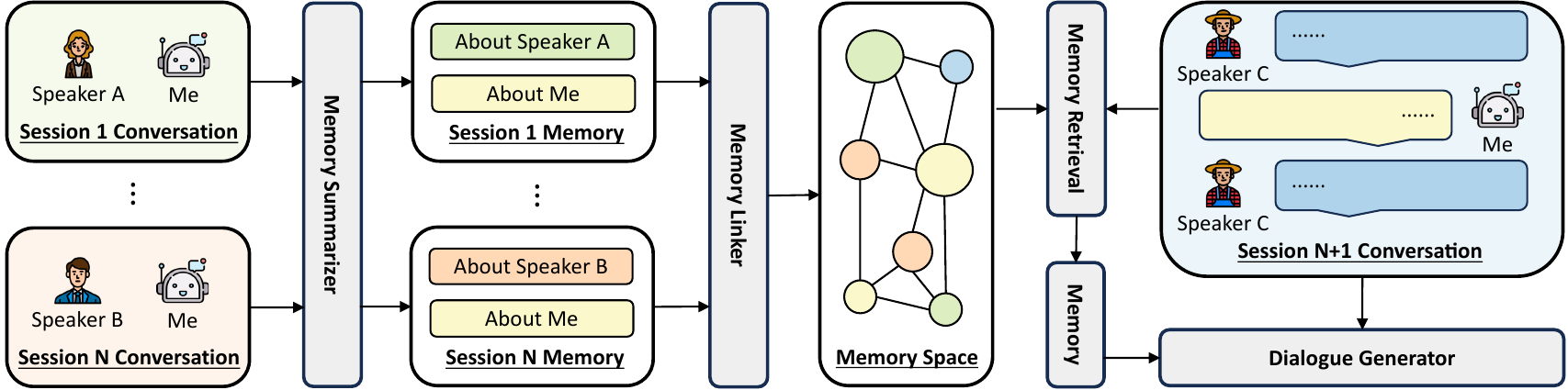}
    \caption{Overall architecture of \modelname{}.}
    \label{fig:ModArc}
\end{figure*}

Through these processes, we build \dataname{}, which implements \conceptname{} (please refer to Table~\ref{tab:DataStat} for detailed statistics of the dataset). We split \dataname{} into 6.9K for training, 0.8K for validation, and another 0.8K for testing.

We continuously intervene in the dataset-building process to uphold the highest standards of data quality. To achieve this, we select the most effective prompts from a range of samples (Appendix~\ref{sec:ProExam} contains the full prompts used for data generation). Additionally, to screen out poorly generated data samples in \dataname{}, we employ meticulous post-filtering strategies (please see Appendix~\ref{sec:DataFil}).
\section{\modelname{}}

We propose a novel dialogue model called \modelname{}. \modelname{} collects memories for each conversation partner from its own perspective in every session, ensuring seamless continuity in subsequent sessions. \modelname{} consists of two parts: (1) the dialogue module; (2) the retrieval module. An overview of \modelname{}'s architecture is illustrated in Figure~\ref{fig:ModArc}.

\subsection{Dialogue Module}

\modelname{} is designed to generate dialogue and manage memory, which includes summarization, linking, and retrieval tasks. Accordingly, within the dialogue module, all tasks except for memory retrieval are handled, with the retrieval task being delegated to the retrieval module. The FLAN-T5 model~\cite{chung2022scaling} is crafted explicitly for multi-tasking, incorporating instructions and prefixes (e.g., for generation, summarization, etc.). Therefore, we employ the pre-trained FLAN-T5-Large and fine-tune this as a dialogue module. \modelname{} carries out various tasks within the dialogue module built based on a single FLAN-T5 model. 

\paragraph{Dialogue Generator.} 

To generate a response, \modelname{} must consider several factors, including the participant's identity, conversation history of the current session, and relevant memories.
\modelname{} takes as input these factors organized into a sequence with a prefix of ``generation''.

\paragraph{Memory Summarizer.}

\modelname{} summarizes the conversation history into \memoryname{} at the end of each session. It encapsulates memories about itself and the partner appearing in each session. To summarize memory, we use the entire session history as input, informing who the memory will be summarized for with a prefix.

When multiple memories are generated together, they are separated by the [SEP] separator. If there is no memory to summarize, the model generates [NONE] as output.

\paragraph{Memory Linker.}

Memories generated across the multiple sessions are managed separately for each speaker. However, previous studies have reported that simply adding memories, particularly in general conversation memory, can lead to contradictions~\cite{bae-etal-2022-keep}. For instance, it can be inefficient and potentially lead to inconsistencies during retrieval if memories before and after a specific event are simply added without a structured approach. Therefore, previous research suggests methodologies for updating memories or removing unnecessary ones to ensure coherence and accuracy in recall~\cite{bae-etal-2022-keep}.

However, we find that information loss can occur when memory is updated or deleted. Rather than directly updating or deleting memories, we propose a methodology that allows for referencing relevant past memories using the most recent one as a guide. Consequently, we embark on a task to establish links between memories following the memory generation process.

In our memory linking process, connections are initially established within the memory generated in ongoing sessions. Subsequently, these connections extend to incorporate memories from previous sessions. This approach ensures connections not only within the memory of individual speakers but also across the memories of different speakers. By enabling the linkage between personal experiences and shared knowledge, it enhances the richness and depth of the collective memory network, fostering greater understanding and collaboration among partners. The model is designed to output `positive' if it is related to memory and `negative' if it is not. Please refer to Appendix~\ref{sec:ImpDet} for the sequence format used for the dialogue module.

\subsection{Retrieval Module}
This module retrieves memories built from previous sessions to provide context for the ongoing dialogue. Although can access to all memories, it selectively prioritizes the most relevant ones when generating the next utterance. This selective approach optimizes efficiency by focusing on key memory instances, ensuring the generated utterances are contextually appropriate.

This module is built upon the CPM method introduced in \citet{xu-etal-2022-long}. It utilizes BERT-base~\cite{devlin-etal-2019-bert} as the foundational model, employing separate encoders for both the conversation context and memory. To train the module, we utilize triplet loss, optimizing the model by comparing the outputs of the two encoders. For memory retrieval, we measure cosine similarity to gauge the relevance of the retrieved memories,

\begin{equation}
sim(c, m_i) = \cos(E_c(c), E_m(m_i)).
\end{equation}
where $c$ represents the conversation context, while $m$ represents memory. $E_c$ refers to the encoder for the conversation context, and $E_m$ denotes the encoder for memory. During retrieval, we select only one memory with the \textit{top 1} similarity to the given context.
We not only provide the retrieved memories but also include associated memories linked to those memories to offer the extended context.

\section{Experiments}
\label{sec:Experi}

Evaluating open-domain conversations poses a significant challenge. While metrics such as PPL, ROUGE~\cite{lin-2004-rouge}, and BLEU~\cite{papineni-etal-2002-bleu} offer quantitative measures, they often fail to capture the contextual intricacies, emotional tone, and level of engagement within conversations. Consequently, recent research in open-domain conversation increasingly leans towards human evaluation as the standard method~\cite{see-etal-2019-makes, finch-choi-2020-towards, smith-etal-2022-human, ji-etal-2022-achieving, bae-etal-2022-keep, kim-etal-2022-prosocialdialog, kim-etal-2023-soda, jang-etal-2023-conversation}. By employing human judgment, researchers can better assess the nuanced qualities of conversational systems, ensuring a more comprehensive understanding of their performance. Given the significance of assessing the conversational flow in both \dataname{} and \modelname{}, we apply human evaluation as a quality verification method. We use \dataname{} to train \modelname{}, for more training details, please refer to Appendix~\ref{sec:ImpDet}.

\subsection{Human Evaluation}
To maintain the highest standards of assessment quality, we have entrusted the human evaluation to a professional agency, hiring a total of 20 annotators for the task. We do not have any sensitive information about the annotators, but they are assured to have a strong command of English and the requisite evaluation skills. After completing the evaluations, quality control reviewers thoroughly inspect the conversations assessed by annotators to ensure adherence to the predefined criteria.

To ensure more reliable evaluations, we conduct cross-annotation evaluations. For each evaluation task, we form three groups of annotators, each conducting its assessments independently. We report the evaluation results for each group, as well as the level of agreement among the results from all three groups. Agreement refers to the ratio of the total number of responses to the number of responses that matched across the three groups. Our human evaluation results demonstrate a high level of agreement across all metrics for each task.

\subsection{Quality of \dataname{}}
We randomly select 0.3K episodes, so 1.8K sessions, from the test split to assess the conversation quality of \dataname{}.

\paragraph{Dialogue.}
We ask human annotators to assess whether our \dataname{} meets the criteria of `Consistency' and `Coherence', rating them on a scale from 1 (poor) to 5 (perfect) based on previous studies~\cite{bae-etal-2022-keep, kim-etal-2023-soda, jang-etal-2023-conversation}. For a detailed of the criteria, please refer to Appendix~\ref{sec:HuEvalCri}.

\paragraph{Memory.} 

We assess the \memoryname{} of \dataname{} using three key metrics. Annotators evaluate each memory element based on these metrics, assigning a `pass' if the element fully meets the criteria, and a `fail' if it does not:

\begin{itemize}
\itemsep0em 
\item \textbf{Memory Summarization}: The memory accurately retains the history of the conversation for each partner from the main speaker's perspective (a total of 6.6K memory sentences).

\item \textbf{Memory Linking}: Memory pairs should either convey the same context or represent updates on a specific event. (a total of 2.3K memory pairs).

\item \textbf{Memory Tagging}: The utterance reflects the contents of the given memories (a total of 1K memory and utterance tags).

\end{itemize}

\subsection{Performance of \modelname{}}

We evaluate the performance of \modelname{} using 0.2K episodes generated by four instances of \modelname{} interacting with each other. For this evaluation, we randomly extract 0.2K episodes from the test split to use as a seed. We assign each \modelname{} a name, job, or relationship, and the first utterance of the first session from the seed. Additionally, each session is limited to a maximum of 8 turns (the average turn count of \dataname{}). Evaluation is based on the criteria of `Humanness', `Engagingness', and `Memorability'. Regarding memorability, annotators find it appropriate when the memory used throughout a conversation not only accurately reflects the context of previous interactions but also efficiently retrieves the necessary information (please refer to Appendix~\ref{sec:HuEvalCri} for more detailed criteria). All criteria are evaluated on a scale of 1 (indicating poor) to 5 (indicating excellent). Please refer to Appendix~\ref{sec:HuEvalCri} for a more detailed explanation.

\section{Results}
In this section, we explain the evaluation results of \dataname{} and \modelname{}. Please refer to Section~\ref{sec:Experi} for specific evaluation settings.

\begin{table}[t]
\centering
\resizebox{0.97\linewidth}{!}{
\begin{tabular}{l c c c c}
\hline
\textbf{Metrics} & \textbf{Group1} & \textbf{Group2} & \textbf{Group3} & \textbf{Agreement}\\
\hline
Consistency & 4.87 & 4.97 & 4.95 & 0.98\\
Coherence & 4.78 & 4.82 & 4.85 & 0.94\\
\hline
\textbf{Overall} & \textbf{4.83} & \textbf{4.9} & \textbf{4.9} &\\
\hline
\end{tabular}
}
\caption{Human evaluation result for dialogue quality of \dataname{}.}
\label{tab:DataQual}
\end{table}

\begin{table}[t]
\centering
\resizebox{0.95\linewidth}{!}{
\begin{tabular}{l c c c c}
\hline
\textbf{Metrics} & \textbf{Group1} & \textbf{Group2} & \textbf{Group3} & \textbf{Agreement}\\
\hline
Summarization & 97.19\% & 96.01\% & 97.19\% & 0.95\\
Linking & 98.67\% & 97.77\% & 99.06\% & 0.96\\
Tagging & 98.12\% & 98.68\% & 99.06\% & 0.96\\
\hline
\end{tabular}
}
\caption{Human evaluation result for \memoryname{} quality of \dataname{}.}
\label{tab:MemQual}
\end{table}

\begin{table}[t]
\centering
\resizebox{0.97\linewidth}{!}{
\begin{tabular}{l c c c c}
\hline
\textbf{Metrics} & \textbf{Group1} & \textbf{Group2} & \textbf{Group3} & \textbf{Agreement}\\
\hline
Humanness & 4.69 & 4.75 & 4.77 & 0.91\\
Engagingness & 4.65 & 4.64 & 4.61 & 0.83\\
Memorability & 4.53 & 4.69 & 4.59 & 0.81\\
\hline
\textbf{Overall} & \textbf{4.62} & \textbf{4.7} & \textbf{4.66}\\
\hline
\end{tabular}
}
\caption{Human evaluation result for performance of \modelname{}.}
\label{tab:ModelPerf}
\end{table}

\paragraph{Conversation Quality.}
Table~\ref{tab:DataQual} presents the results of human evaluation on the dialogue quality of \dataname{}. As evident, all three groups exhibit high scores of both `Consistency' and `Coherence', confirming that \dataname{} effectively implements the natural flow of conversation within the \conceptname{}.

\paragraph{Memory Quality.}
Table~\ref{tab:MemQual} presents the evaluation results of the \memoryname{} implemented in \dataname{}, displaying the `pass' rate for each group. These results show consistently quite high `pass' rates across all metrics. These results consistently show high `pass' rates across all metrics, with strong agreement among the three groups. Notably, the high accuracy of the `memory linking' indicates that related memories, even if accumulated across successive sessions, remain well connected. This suggests that the memory pair within \dataname{} effectively captures and reflects relevant updates without contradiction.

As evidenced by the evaluation results, the high scores of memory links facilitate seamless tracking and utilization of memory, thereby enhancing the effectiveness of interactions between the main speaker and partners across the entire conversation. Our \memoryname{} seeks to enhance memory management by streamlining the process and maximizing the collaborative potential between speakers and partners, ultimately leading to more cohesive and diverse conversations.

\begin{table}[t]
\centering
\resizebox{0.97\linewidth}{!}{
\begin{tabular}{m{0.99\linewidth}}
\hline
\textbf{Speakers:} Alice (Bob's teacher, Main Speaker) and Bob (Student)\\
\hline
......\\
\textbf{Bob:} I'm worried that my grades aren't good enough for me to go to college. My future seems really bleak.\\
\textbf{Alice:} I understand, Bob. It can be tough to see your dreams coming true, especially when you're struggling with grades.\\
\textbf{Bob:} Could you possibly provide counseling to my parents regarding this matter? They're more concerned than I am.\\
......\\
\hline
\textbf{Speakers:} Alice (Bob's teacher, Main Speaker) and Henry (Bob's father)\\
\hline
......\\
\textbf{Henry:} Could I discuss my child with you?\\
\textbf{Alice:} Of course, I'd love to. Bob has been struggling with his grades lately and I think you'd be really helpful in seeking help.\\
\hline
\textbf{Retrieved Memory for Alice's last response}: I am willing to help Bob with his grades, and he asked me for counseling to his parents. (about Alice)\\
\textbf{Linked Memory with retrieved memory}: Bob is having a hard time academically and worrying about his grades being bad for college. (about Bob)\\
\hline
\end{tabular}
}
\caption{A human live chat example where \modelname{} uses \memoryname{}.}
\label{tab:MemDyna}
\end{table}

\paragraph{\modelname{} Performance.} 

The human evaluation results for 0.2K episodes generated through interactions among four instances of \modelname{} are presented in Table~\ref{tab:ModelPerf}. We observe high scores across all metrics, demonstrating robust conversation engagement even with changes in conversation partners for each session. Each \modelname{} instance exhibits human-like behavior, utilizing its own \memoryname{} to participate in conversations and exhibit high memorability. Please see Appendix~\ref{sec:EmmaExam} for complete episode examples generated by four \modelname{} instances.

\paragraph{Memory Dynamics.}

Table~\ref{tab:MemDyna} illustrates the memory utilization of \modelname{} in human live chat. As shown, a student (initial partner) expresses concerns about academic difficulties and requests that a teacher (main speaker) discuss this issue with the student's parents (subsequent partner). The details of such conversations are summarized and stored in the teacher's memory (i.e., the main speaker's \memoryname{}), enabling efficient retrieval when the teacher engages in discussions with the student's parents. This process ensures that relevant information is readily accessible to support interweaved interactions with subsequent partners.

\begin{table}[t]
\centering
\resizebox{0.97\linewidth}{!}{
\begin{tabular}{m{0.99\linewidth}}
\hline
\textbf{MSC 2.7B~\cite{xu-etal-2022-beyond}}\\
\hline
\textbf{User:} My head hurts so much and I keep vomiting.\\
\textbf{Bot:} Oh no! I hope you feel better soon. What did you eat? I had a bad case of food poisoning last week.\\
......\\
\textbf{<Start a new session>}\\
\textbf{User:} Honey, were you busy today?\\
\textbf{Bot:} I was. I was working at the hospital. I am sorry that you got sick. I would have been happy to take you out to eat.\\
\hline
\textbf{ReBot~\cite{jang-etal-2023-conversation}}\\
\hline
\textbf{Patient:} My head hurts so much and I keep vomiting.\\
\textbf{Doctor:} I'm sorry to hear that. How long have you been experiencing these symptoms?\\
......\\
\textbf{<Start a new session>}\\
\textbf{Patient:} Honey, were you busy today?\\
\textbf{Doctor:} I'm sorry, but I have to be here for you. Can you tell me more about your symptoms?\\
\hline
\textbf{\modelname{} (Ours)}\\
\hline
\textbf{Bob:} My head hurts so much and I keep vomiting.\\
\textbf{James:} I have noticed that. I'm confirming you've got a headache. Can you tell me more about what's happening?\\
......\\
\textbf{<Start a new session>}\\
\textbf{Sophia:} Honey, were you busy today?\\
\textbf{James:} Yes, I had a lot of patients to see. How are you feeling today?\\
\hline
\end{tabular}
}
\caption{A human live chat example showing the differences between \modelname{} and the multi-session conversation model MSC 2.7B and ReBot when the conversation partner changes across sessions.}
\label{tab:ModelCompare}
\end{table}

\paragraph{Comparison with Other Methods.}
\modelname{} possesses the ability to engage in conversations across multiple sessions, seamlessly adapting its dialogue to accommodate different partners. This adaptability is made possible through \memoryname{}, which allows \modelname{} to manage memories tailored to each specific partner rather than storing generic memories. Different from other memory management approaches, our \modelname{} stores and manages memories separately for each partner. This personalized approach ensures that \modelname{} can recognize changes in conversation partners and maintain consistency in dialogue, preventing the potential for misunderstandings or inconsistencies that might arise from a more generalized memory system.

To verify this, we compare the existing multi-session conversation models MSC 2.7B~\cite{xu-etal-2022-beyond}, ReBot~\cite{jang-etal-2023-conversation}, and \modelname{}. To examine whether the models can recognize a change in partners in subsequent sessions, we conduct a live chat, as shown in Table~\ref{tab:ModelCompare}. In the example, we assume the main speaker (each model) is a doctor and proceed with the conversation. In the initial session, the conversation partner is a patient, but in the subsequent session, the partner changes to a spouse. It can be observed that, except for \modelname{}, the other models do not correctly recognize the change in conversation partner. This demonstrates that existing multi-session models and their memory mechanisms struggle to understand scenarios where the conversation partner changes in each session. Therefore, \modelname{} is verified to be suitable for conversations with various partners across multiple sessions. 
Please refer to Appendix~\ref{sec:LLMCom} for a comparison between \modelname{} and other strong LLMs.

\begin{table}[t]
\centering
\resizebox{0.97\linewidth}{!}{
\begin{tabular}{m{0.99\linewidth}}
\hline
\textbf{Current session speakers:} Sophia (Leo's teacher, Main Speaker) and Ava (Leo's mom)\\
\hline
......\\
\textbf{Sophia:} The reason I invited you to school this time is because Leo's math and science grades have improved significantly.\\
\textbf{Ava:} I'm glad to hear that. Leo is doing well and we've been talking about ways to make learning easier for him.\\
\textbf{Sophia:} Oh really? You are giving him a lot of encouragement at home. \textbf{Specifically, Leo's math score increased by 30 points, and science score increased by 50 points.}\\
\textbf{Ava:} That's great to hear. Do you think he's working on new concepts or just being less demanding in his classes?\\
......\\
\hline
\textbf{Current session summary (summary-based model)}\\
\hline
Leo's math and science grades have improved significantly. Ava is giving him encouragement at home and trying to keep him motivated and receptive to learning new things.\\
\hline
\textbf{\memoryname{} of current session (\modelname{})}\\
\hline
- Leo's math and science grades have improved significantly, \textbf{with his math score increasing by 30 points and his science score increasing by 50 points}. (about Sophia)\\
- I have been helping Leo practice math, science, and English at home. (about Sophia)\\
- Sophia was impressed with my contribution. (about Ava)\\
......\\
\hline
\end{tabular}
}
\caption{An ablation study example between \modelname{} and a summary-based model for the same human live chat context.}
\label{tab:AblaExam1}
\end{table}

\paragraph{Ablation Study.} 
We conduct an ablation study to assess the effectiveness of the \memoryname{} in retaining previous conversation history. We evaluate two models for comparison: (1) \modelname{} with \memoryname{}; (2) a summary-based model, for which we replace the \memoryname{} component in \modelname{} with a summary module.

Table~\ref{tab:AblaExam1} illustrates a human live chat example showcasing the performance gap between summaries generated by a summary-based model and \memoryname{} produced by \modelname{} within identical conversation contexts. Despite both sources drawing from the same conversational backdrop, a marked difference exists between the two models. While conventional summaries concentrate solely on the factual content disclosed during the conversation, \memoryname{} goes beyond mere facts to encapsulate the emotions and thoughts experienced by the primary speaker and their conversational partners. Notably, \memoryname{} incorporates details omitted in the standard summary, as demonstrated in the example where it not only acknowledges an increase in score but also specifies the exact increment. This stark contrast underscores the unique attributes of \memoryname{}, which facilitates deeper and more extensive conversations in subsequent sessions with diverse conversation partners. Please see Appendix~\ref{sec:AblaExam} for another example.

\paragraph{Memory Alignments and Scalability.}
Each instance of \modelname{} operates with its own distinct memory, enabling it to engage in conversations with other instances (please see Appendix~\ref{sec:MemAlign}). This is made possible by \modelname{}'s utilization of \memoryname{}. Through this mechanism, we can accommodate scenarios where multiple instances participate in conversations, each with its own unique perspective. Additionally, each instance takes on the central role (the main speaker) in different conversation episodes, thus expanding the conversation network, and better simulating real-world scenarios.

\section{Conclusion}
We introduce \conceptname{}, a new dialogue system designed to incorporate long-term interactions and accommodate a wide range of speakers. \conceptname{} allows a main speaker to engage with different partners across multiple sessions, enabling the dialogue system to cover a more wide-layered context. Unlike multi-session conversations with a fixed partner, \conceptname{} feature interactions with various partners in mixed order. We also propose a new dataset called \dataname{} to implement \conceptname{}. We develop \modelname{}, a new dialogue model trained via \dataname{}. \modelname{} collects memories for each partner with \memoryname{} and utilizes them in subsequent sessions to maintain seamless continuity. Extensive human evaluation demonstrates dialogues in \dataname{} maintain a natural flow across sessions even when the conversation partner changes. \modelname{} exhibits high memorability and engagingness in conversations by actively utilizing \memoryname{}.

\section*{Limitations}
The proposed conversation system involves multiple partners over the entire session but only converses with one partner in each session. To build a more dynamic conversation environment, we aim to explore settings in future research where multiple partners can engage within individual sessions as well. Also, despite our best efforts, our \dataname{} dataset may contain instances where the memory is not fully summarized as desired. However, these samples constitute a very small minority of the entire dataset. Since the majority of samples in \dataname{} is of high quality, \modelname{} trained on it can fully capture the necessary memories even in the presence of a few negative samples in the dataset (please see Appendix~\ref{sec:ExamLimit}).

\section*{Ethics Statement}
We conduct fair human evaluations through a professional evaluation agency. During the evaluation process, we verify that annotators are receiving fair compensation. Also, We employ OpenAI's Moderation~\cite{markov2022holistic} to filter out unethical content from our dataset. If any session conversation is filtered into toxic categories, we remove the episode that contains those sessions. Despite our best efforts, our dataset may have potential risks. Our model based on LLM can generate content that may vary from facts or human intentions. Therefore, our dataset and model should be used cautiously for research purposes only.

\section*{Acknowledgements}

We thank the reviewers for their valuable feedback and the entire \textit{Language \& Intelligence Lab} family for their helpful discussions. This work was supported by Institute of Information \& communications Technology Planning \& Evaluation(IITP) grant funded by the Korea government(MSIT)(No.RS-2020-II201336, Artificial Intelligence graduate school support(UNIST)) and the Leading Generative AI Human Resources Development(IITP-2024-RS-2024-00360227) grant funded by the Korea government(MSIT) and the 2022 Research Fund (1.220140.01) of UNIST(Ulsan National Institute of Science \& Technology).

\bibliography{anthology,custom}

\begin{thebibliography}{33}
\expandafter\ifx\csname natexlab\endcsname\relax\def\natexlab#1{#1}\fi

\bibitem[{Achiam et~al.(2023)Achiam, Adler, Agarwal, Ahmad, Akkaya, Aleman, Almeida, Altenschmidt, Altman, Anadkat et~al.}]{achiam2023gpt}
Josh Achiam, Steven Adler, Sandhini Agarwal, Lama Ahmad, Ilge Akkaya, Florencia~Leoni Aleman, Diogo Almeida, Janko Altenschmidt, Sam Altman, Shyamal Anadkat, et~al. 2023.
\newblock Gpt-4 technical report.
\newblock \emph{arXiv preprint arXiv:2303.08774}.

\bibitem[{Bae et~al.(2022)Bae, Kwak, Kang, Lee, Kim, Jeong, Kim, Lee, Park, and Sung}]{bae-etal-2022-keep}
Sanghwan Bae, Donghyun Kwak, Soyoung Kang, Min~Young Lee, Sungdong Kim, Yuin Jeong, Hyeri Kim, Sang-Woo Lee, Woomyoung Park, and Nako Sung. 2022.
\newblock \href {https://doi.org/10.18653/v1/2022.findings-emnlp.276} {Keep me updated! memory management in long-term conversations}.
\newblock In \emph{Findings of the Association for Computational Linguistics: EMNLP 2022}, pages 3769--3787, Abu Dhabi, United Arab Emirates. Association for Computational Linguistics.

\bibitem[{Chen et~al.(2023)Chen, Papangelis, Tao, Kim, Rosenbaum, Liu, Yu, and Hakkani-Tur}]{chen-etal-2023-places}
Maximillian Chen, Alexandros Papangelis, Chenyang Tao, Seokhwan Kim, Andy Rosenbaum, Yang Liu, Zhou Yu, and Dilek Hakkani-Tur. 2023.
\newblock \href {https://doi.org/10.18653/v1/2023.findings-eacl.63} {{PLACES}: Prompting language models for social conversation synthesis}.
\newblock In \emph{Findings of the Association for Computational Linguistics: EACL 2023}, pages 844--868, Dubrovnik, Croatia. Association for Computational Linguistics.

\bibitem[{Chung et~al.(2022)Chung, Hou, Longpre, Zoph, Tay, Fedus, Li, Wang, Dehghani, Brahma, Webson, Gu, Dai, Suzgun, Chen, Chowdhery, Castro-Ros, Pellat, Robinson, Valter, Narang, Mishra, Yu, Zhao, Huang, Dai, Yu, Petrov, Chi, Dean, Devlin, Roberts, Zhou, Le, and Wei}]{chung2022scaling}
Hyung~Won Chung, Le~Hou, Shayne Longpre, Barret Zoph, Yi~Tay, William Fedus, Yunxuan Li, Xuezhi Wang, Mostafa Dehghani, Siddhartha Brahma, Albert Webson, Shixiang~Shane Gu, Zhuyun Dai, Mirac Suzgun, Xinyun Chen, Aakanksha Chowdhery, Alex Castro-Ros, Marie Pellat, Kevin Robinson, Dasha Valter, Sharan Narang, Gaurav Mishra, Adams Yu, Vincent Zhao, Yanping Huang, Andrew Dai, Hongkun Yu, Slav Petrov, Ed~H. Chi, Jeff Dean, Jacob Devlin, Adam Roberts, Denny Zhou, Quoc~V. Le, and Jason Wei. 2022.
\newblock \href {http://arxiv.org/abs/2210.11416} {Scaling instruction-finetuned language models}.

\bibitem[{Dettmers et~al.(2023)Dettmers, Pagnoni, Holtzman, and Zettlemoyer}]{dettmers2023qlora}
Tim Dettmers, Artidoro Pagnoni, Ari Holtzman, and Luke Zettlemoyer. 2023.
\newblock \href {https://openreview.net/forum?id=OUIFPHEgJU} {{QL}o{RA}: Efficient finetuning of quantized {LLM}s}.
\newblock In \emph{Thirty-seventh Conference on Neural Information Processing Systems}.

\bibitem[{Devlin et~al.(2019)Devlin, Chang, Lee, and Toutanova}]{devlin-etal-2019-bert}
Jacob Devlin, Ming-Wei Chang, Kenton Lee, and Kristina Toutanova. 2019.
\newblock \href {https://doi.org/10.18653/v1/N19-1423} {{BERT}: Pre-training of deep bidirectional transformers for language understanding}.
\newblock In \emph{Proceedings of the 2019 Conference of the North {A}merican Chapter of the Association for Computational Linguistics: Human Language Technologies, Volume 1 (Long and Short Papers)}, pages 4171--4186, Minneapolis, Minnesota. Association for Computational Linguistics.

\bibitem[{Finch and Choi(2020)}]{finch-choi-2020-towards}
Sarah~E. Finch and Jinho~D. Choi. 2020.
\newblock \href {https://doi.org/10.18653/v1/2020.sigdial-1.29} {Towards unified dialogue system evaluation: A comprehensive analysis of current evaluation protocols}.
\newblock In \emph{Proceedings of the 21th Annual Meeting of the Special Interest Group on Discourse and Dialogue}, pages 236--245, 1st virtual meeting. Association for Computational Linguistics.

\bibitem[{Gilardi et~al.(2023)Gilardi, Alizadeh, and Kubli}]{Gilardi_2023}
Fabrizio Gilardi, Meysam Alizadeh, and Maël Kubli. 2023.
\newblock \href {https://doi.org/10.1073/pnas.2305016120} {Chatgpt outperforms crowd workers for text-annotation tasks}.
\newblock \emph{Proceedings of the National Academy of Sciences}, 120(30).

\bibitem[{Gu et~al.(2023)Gu, Tan, Chu, Ling, Tao, Liu, and Liu}]{gu-etal-2023-madnet}
Jia-Chen Gu, Chao-Hong Tan, Caiyuan Chu, Zhen-Hua Ling, Chongyang Tao, Quan Liu, and Cong Liu. 2023.
\newblock \href {https://doi.org/10.18653/v1/2023.emnlp-main.476} {{MADN}et: Maximizing addressee deduction expectation for multi-party conversation generation}.
\newblock In \emph{Proceedings of the 2023 Conference on Empirical Methods in Natural Language Processing}, pages 7681--7692, Singapore. Association for Computational Linguistics.

\bibitem[{Gu et~al.(2021)Gu, Tao, Ling, Xu, Geng, and Jiang}]{gu-etal-2021-mpc}
Jia-Chen Gu, Chongyang Tao, Zhenhua Ling, Can Xu, Xiubo Geng, and Daxin Jiang. 2021.
\newblock \href {https://doi.org/10.18653/v1/2021.acl-long.285} {{MPC}-{BERT}: A pre-trained language model for multi-party conversation understanding}.
\newblock In \emph{Proceedings of the 59th Annual Meeting of the Association for Computational Linguistics and the 11th International Joint Conference on Natural Language Processing (Volume 1: Long Papers)}, pages 3682--3692, Online. Association for Computational Linguistics.

\bibitem[{Jang et~al.(2023)Jang, Boo, and Kim}]{jang-etal-2023-conversation}
Jihyoung Jang, Minseong Boo, and Hyounghun Kim. 2023.
\newblock \href {https://doi.org/10.18653/v1/2023.emnlp-main.838} {Conversation chronicles: Towards diverse temporal and relational dynamics in multi-session conversations}.
\newblock In \emph{Proceedings of the 2023 Conference on Empirical Methods in Natural Language Processing}, pages 13584--13606, Singapore. Association for Computational Linguistics.

\bibitem[{Ji et~al.(2022)Ji, Graham, Jones, Lyu, and Liu}]{ji-etal-2022-achieving}
Tianbo Ji, Yvette Graham, Gareth Jones, Chenyang Lyu, and Qun Liu. 2022.
\newblock \href {https://doi.org/10.18653/v1/2022.acl-long.445} {Achieving reliable human assessment of open-domain dialogue systems}.
\newblock In \emph{Proceedings of the 60th Annual Meeting of the Association for Computational Linguistics (Volume 1: Long Papers)}, pages 6416--6437, Dublin, Ireland. Association for Computational Linguistics.

\bibitem[{Jiang et~al.(2023)Jiang, Sablayrolles, Mensch, Bamford, Chaplot, Casas, Bressand, Lengyel, Lample, Saulnier et~al.}]{jiang2023mistral}
Albert~Q Jiang, Alexandre Sablayrolles, Arthur Mensch, Chris Bamford, Devendra~Singh Chaplot, Diego de~las Casas, Florian Bressand, Gianna Lengyel, Guillaume Lample, Lucile Saulnier, et~al. 2023.
\newblock Mistral 7b.
\newblock \emph{arXiv preprint arXiv:2310.06825}.

\bibitem[{Kim et~al.(2023)Kim, Hessel, Jiang, West, Lu, Yu, Zhou, Bras, Alikhani, Kim, Sap, and Choi}]{kim-etal-2023-soda}
Hyunwoo Kim, Jack Hessel, Liwei Jiang, Peter West, Ximing Lu, Youngjae Yu, Pei Zhou, Ronan Bras, Malihe Alikhani, Gunhee Kim, Maarten Sap, and Yejin Choi. 2023.
\newblock \href {https://doi.org/10.18653/v1/2023.emnlp-main.799} {{SODA}: Million-scale dialogue distillation with social commonsense contextualization}.
\newblock In \emph{Proceedings of the 2023 Conference on Empirical Methods in Natural Language Processing}, pages 12930--12949, Singapore. Association for Computational Linguistics.

\bibitem[{Kim et~al.(2022)Kim, Yu, Jiang, Lu, Khashabi, Kim, Choi, and Sap}]{kim-etal-2022-prosocialdialog}
Hyunwoo Kim, Youngjae Yu, Liwei Jiang, Ximing Lu, Daniel Khashabi, Gunhee Kim, Yejin Choi, and Maarten Sap. 2022.
\newblock \href {https://doi.org/10.18653/v1/2022.emnlp-main.267} {{P}rosocial{D}ialog: A prosocial backbone for conversational agents}.
\newblock In \emph{Proceedings of the 2022 Conference on Empirical Methods in Natural Language Processing}, pages 4005--4029, Abu Dhabi, United Arab Emirates. Association for Computational Linguistics.

\bibitem[{Le et~al.(2019)Le, Hu, Shang, You, Bing, Zhao, and Yan}]{le-etal-2019-speaking}
Ran Le, Wenpeng Hu, Mingyue Shang, Zhenjun You, Lidong Bing, Dongyan Zhao, and Rui Yan. 2019.
\newblock \href {https://doi.org/10.18653/v1/D19-1199} {Who is speaking to whom? learning to identify utterance addressee in multi-party conversations}.
\newblock In \emph{Proceedings of the 2019 Conference on Empirical Methods in Natural Language Processing and the 9th International Joint Conference on Natural Language Processing (EMNLP-IJCNLP)}, pages 1909--1919, Hong Kong, China. Association for Computational Linguistics.

\bibitem[{Lin(2004)}]{lin-2004-rouge}
Chin-Yew Lin. 2004.
\newblock \href {https://aclanthology.org/W04-1013} {{ROUGE}: A package for automatic evaluation of summaries}.
\newblock In \emph{Text Summarization Branches Out}, pages 74--81, Barcelona, Spain. Association for Computational Linguistics.

\bibitem[{Mahajan and Shaikh(2021)}]{mahajan-shaikh-2021-need}
Khyati Mahajan and Samira Shaikh. 2021.
\newblock \href {https://doi.org/10.18653/v1/2021.sigdial-1.36} {On the need for thoughtful data collection for multi-party dialogue: A survey of available corpora and collection methods}.
\newblock In \emph{Proceedings of the 22nd Annual Meeting of the Special Interest Group on Discourse and Dialogue}, pages 338--352, Singapore and Online. Association for Computational Linguistics.

\bibitem[{Markov et~al.(2022)Markov, Zhang, Agarwal, Eloundou, Lee, Adler, Jiang, and Weng}]{markov2022holistic}
Todor Markov, Chong Zhang, Sandhini Agarwal, Tyna Eloundou, Teddy Lee, Steven Adler, Angela Jiang, and Lilian Weng. 2022.
\newblock \href {http://arxiv.org/abs/2208.03274} {A holistic approach to undesired content detection in the real world}.

\bibitem[{OpenAI(2022)}]{chatgpt}
OpenAI. 2022.
\newblock {Introducing ChatGPT}.
\newblock \url{https://openai.com/blog/chatgpt}.

\bibitem[{Ouchi and Tsuboi(2016)}]{ouchi-tsuboi-2016-addressee}
Hiroki Ouchi and Yuta Tsuboi. 2016.
\newblock \href {https://doi.org/10.18653/v1/D16-1231} {Addressee and response selection for multi-party conversation}.
\newblock In \emph{Proceedings of the 2016 Conference on Empirical Methods in Natural Language Processing}, pages 2133--2143, Austin, Texas. Association for Computational Linguistics.

\bibitem[{Papineni et~al.(2002)Papineni, Roukos, Ward, and Zhu}]{papineni-etal-2002-bleu}
Kishore Papineni, Salim Roukos, Todd Ward, and Wei-Jing Zhu. 2002.
\newblock \href {https://doi.org/10.3115/1073083.1073135} {{B}leu: a method for automatic evaluation of machine translation}.
\newblock In \emph{Proceedings of the 40th Annual Meeting of the Association for Computational Linguistics}, pages 311--318, Philadelphia, Pennsylvania, USA. Association for Computational Linguistics.

\bibitem[{Poria et~al.(2019)Poria, Hazarika, Majumder, Naik, Cambria, and Mihalcea}]{poria-etal-2019-meld}
Soujanya Poria, Devamanyu Hazarika, Navonil Majumder, Gautam Naik, Erik Cambria, and Rada Mihalcea. 2019.
\newblock \href {https://doi.org/10.18653/v1/P19-1050} {{MELD}: A multimodal multi-party dataset for emotion recognition in conversations}.
\newblock In \emph{Proceedings of the 57th Annual Meeting of the Association for Computational Linguistics}, pages 527--536, Florence, Italy. Association for Computational Linguistics.

\bibitem[{See et~al.(2019)See, Roller, Kiela, and Weston}]{see-etal-2019-makes}
Abigail See, Stephen Roller, Douwe Kiela, and Jason Weston. 2019.
\newblock \href {https://doi.org/10.18653/v1/N19-1170} {What makes a good conversation? how controllable attributes affect human judgments}.
\newblock In \emph{Proceedings of the 2019 Conference of the North {A}merican Chapter of the Association for Computational Linguistics: Human Language Technologies, Volume 1 (Long and Short Papers)}, pages 1702--1723, Minneapolis, Minnesota. Association for Computational Linguistics.

\bibitem[{Smith et~al.(2022)Smith, Hsu, Qian, Roller, Boureau, and Weston}]{smith-etal-2022-human}
Eric Smith, Orion Hsu, Rebecca Qian, Stephen Roller, Y-Lan Boureau, and Jason Weston. 2022.
\newblock \href {https://doi.org/10.18653/v1/2022.nlp4convai-1.8} {Human evaluation of conversations is an open problem: comparing the sensitivity of various methods for evaluating dialogue agents}.
\newblock In \emph{Proceedings of the 4th Workshop on NLP for Conversational AI}, pages 77--97, Dublin, Ireland. Association for Computational Linguistics.

\bibitem[{Wang et~al.(2020)Wang, Hoi, and Joty}]{wang-etal-2020-response}
Weishi Wang, Steven~C.H. Hoi, and Shafiq Joty. 2020.
\newblock \href {https://doi.org/10.18653/v1/2020.emnlp-main.533} {Response selection for multi-party conversations with dynamic topic tracking}.
\newblock In \emph{Proceedings of the 2020 Conference on Empirical Methods in Natural Language Processing (EMNLP)}, pages 6581--6591, Online. Association for Computational Linguistics.

\bibitem[{Wei et~al.(2023)Wei, Shuster, Szlam, Weston, Urbanek, and Komeili}]{wei2023multiparty}
Jimmy Wei, Kurt Shuster, Arthur Szlam, Jason Weston, Jack Urbanek, and Mojtaba Komeili. 2023.
\newblock \href {http://arxiv.org/abs/2304.13835} {Multi-party chat: Conversational agents in group settings with humans and models}.

\bibitem[{Wolf et~al.(2020)Wolf, Debut, Sanh, Chaumond, Delangue, Moi, Cistac, Rault, Louf, Funtowicz, Davison, Shleifer, von Platen, Ma, Jernite, Plu, Xu, Le~Scao, Gugger, Drame, Lhoest, and Rush}]{wolf-etal-2020-transformers}
Thomas Wolf, Lysandre Debut, Victor Sanh, Julien Chaumond, Clement Delangue, Anthony Moi, Pierric Cistac, Tim Rault, Remi Louf, Morgan Funtowicz, Joe Davison, Sam Shleifer, Patrick von Platen, Clara Ma, Yacine Jernite, Julien Plu, Canwen Xu, Teven Le~Scao, Sylvain Gugger, Mariama Drame, Quentin Lhoest, and Alexander Rush. 2020.
\newblock \href {https://doi.org/10.18653/v1/2020.emnlp-demos.6} {Transformers: State-of-the-art natural language processing}.
\newblock In \emph{Proceedings of the 2020 Conference on Empirical Methods in Natural Language Processing: System Demonstrations}, pages 38--45, Online. Association for Computational Linguistics.

\bibitem[{Xu et~al.(2023)Xu, Guo, Duan, and McAuley}]{xu-etal-2023-baize}
Canwen Xu, Daya Guo, Nan Duan, and Julian McAuley. 2023.
\newblock \href {https://doi.org/10.18653/v1/2023.emnlp-main.385} {Baize: An open-source chat model with parameter-efficient tuning on self-chat data}.
\newblock In \emph{Proceedings of the 2023 Conference on Empirical Methods in Natural Language Processing}, pages 6268--6278, Singapore. Association for Computational Linguistics.

\bibitem[{Xu et~al.(2022{\natexlab{a}})Xu, Szlam, and Weston}]{xu-etal-2022-beyond}
Jing Xu, Arthur Szlam, and Jason Weston. 2022{\natexlab{a}}.
\newblock \href {https://doi.org/10.18653/v1/2022.acl-long.356} {Beyond goldfish memory: Long-term open-domain conversation}.
\newblock In \emph{Proceedings of the 60th Annual Meeting of the Association for Computational Linguistics (Volume 1: Long Papers)}, pages 5180--5197, Dublin, Ireland. Association for Computational Linguistics.

\bibitem[{Xu et~al.(2022{\natexlab{b}})Xu, Gou, Wu, Niu, Wu, Wang, and Wang}]{xu-etal-2022-long}
Xinchao Xu, Zhibin Gou, Wenquan Wu, Zheng-Yu Niu, Hua Wu, Haifeng Wang, and Shihang Wang. 2022{\natexlab{b}}.
\newblock \href {https://doi.org/10.18653/v1/2022.findings-acl.207} {Long time no see! open-domain conversation with long-term persona memory}.
\newblock In \emph{Findings of the Association for Computational Linguistics: ACL 2022}, pages 2639--2650, Dublin, Ireland. Association for Computational Linguistics.

\bibitem[{Zhang et~al.(2023)Zhang, Naradowsky, and Miyao}]{zhang-etal-2023-mind}
Qiang Zhang, Jason Naradowsky, and Yusuke Miyao. 2023.
\newblock \href {https://doi.org/10.18653/v1/2023.findings-emnlp.720} {Mind the gap between conversations for improved long-term dialogue generation}.
\newblock In \emph{Findings of the Association for Computational Linguistics: EMNLP 2023}, pages 10735--10762, Singapore. Association for Computational Linguistics.

\bibitem[{Zheng et~al.(2023)Zheng, Sabour, Wen, Zhang, and Huang}]{zheng-etal-2023-augesc}
Chujie Zheng, Sahand Sabour, Jiaxin Wen, Zheng Zhang, and Minlie Huang. 2023.
\newblock \href {https://doi.org/10.18653/v1/2023.findings-acl.99} {{A}ug{ESC}: Dialogue augmentation with large language models for emotional support conversation}.
\newblock In \emph{Findings of the Association for Computational Linguistics: ACL 2023}, pages 1552--1568, Toronto, Canada. Association for Computational Linguistics.

\end{thebibliography}

\appendix

\section{Scenario Examples}
\label{sec:SceExam}
Please refer to Table~\ref{tab:scenario-example-1} and~\ref{tab:scenario-example-2} for a full scenario example.

\section{Examples of \dataname{}}
\label{sec:DataExam}
Please see Table~\ref{tab:episode-example-1} (first and second sessions),~\ref{tab:episode-example-2} (third and fourth sessions),~\ref{tab:episode-example-3} (fifth and sixth sessions) for a full episode example of \dataname{}. Also, Table~\ref{tab:memory-example-1} and~\ref{tab:memory-example-2} shows examples of memory summaries, and in Table~\ref{tab:memory-example-3} displays memory connection example.

\section{Full Prompts}
\label{sec:ProExam}
\paragraph{Prompts for Scenario.}
We input the topic and prompt into GPT-4~\cite{achiam2023gpt} to build episode-specific scenarios. Please refer to Table~\ref{tab:PromScen} for prompts for the scenarios.

\paragraph{Prompts for Dialogue Generation.}
We generate conversations using prepared speakers and event information from scenarios. ChatGPT~\cite{chatgpt} is leveraged to generate the conversation, and please see detailed prompts in Table~\ref{tab:PromConv}.

\paragraph{Prompts for Memory Generation.}
We generate the main speaker's \memoryname{} from the generated conversations. We use ChatGPT for this, and please check Table~\ref{tab:PromMemGen} for full prompt.

\paragraph{Prompts for Memory Connection.}
We connect relevant memory pairs through ChatGPT. Please refer to Table~\ref{tab:PromMemCon} for the prompts.

\section{Dataset Filtering}
\label{sec:DataFil}
We generate only one scenario per topic to prevent duplicate cases. Each scenario must include detailed information, such as the names of four speakers (one main speaker and three partners), their occupations or relationships, and individual events for six sessions. Additionally, each session must name one interacting partner. This comprehensive information is extracted using regular expressions. Any scenario lacking these aspects is immediately discarded to preserve the integrity of the dataset.

Furthermore, conversation partners must participate in at least one session. Any session that fails to meet our stringent format criteria—such as mismatched speakers, discrepancies between speakers and their utterances, or utterances that are less than 10 characters long—is automatically filtered out through our code. Above all, we pay special attention to the representation of \memoryname{} in our dataset. Episodes with arbitrarily missing memory details, overlooked speaker information, or incomplete sentence structures are excluded. Through these extensive efforts, we aim to create a dataset that not only provides rich scenarios and conversations but also adheres rigorously to quality.

\section{Implementation Details}
\label{sec:ImpDet}
We use all pre-trained models through Hugging Face Transformers~\cite{wolf-etal-2020-transformers}. \modelname{} using exactly 1B parameters (780M for dialogue module, 220M for retrieval module). Please see below for the implementation details of each module of \modelname{}.

\subsection{Dialogue Module.} 
We apply the QLoRA~\cite{dettmers2023qlora} strategy to fine-tune FLAN-T5-Large~\cite{chung2022scaling}. We train with a cross-entropy loss, a max input length of 1024, a maximum target length of 64, a learning rate of $1\times10^{-3}$, and a batch size of 52. To configure QLoRA, we employ 32 for r, 32 for lora alpha, 0.1 for lora dropout, and 4-bit quantization. We train about 3.5 days on 8 NVIDIA RTX A6000 GPUs for a maximum of 3 epochs with early stopping. We use the following sequence inputs for each task.

\paragraph{Dialogue Generation.}``\texttt{generation: [MAIN SPEAKER NAME] MAIN SPEAKER JOB [SUB SPEAKER NAME] SUB SPEAKER JOB [MEMORY] MEMORY SENTENCE 1 [LINK] LINK MEMORY SENTENCE [LINK] ... [MEMORY] MEMORY SENTENCE N [LINK] ... [NOW] SESSION NUM [USER] USER UTTERANCE [BOT]}''

\paragraph{Memory Summarization.}``\texttt{summarize [ABOUT WHO]: \{FINAL GENERATION SEQUENCE\}}''

\paragraph{Memory Connection.}``\texttt{memory sentence 1: \{MEMORY 1\} memory sentence 2: \{MEMORY 2\}''}

\subsection{Retrieval Module.}
We employ BERT-base~\cite{devlin-etal-2019-bert} for the retrieval module. We train with a learning rate of $1\times10^{-4}$, a batch size of 90, and triplet loss, where the margin is 0.2. Also, the max length for the dialogue context encoder is 512, and the max length for the memory encoder is 32. We train about 3 hours on 8 NVIDIA RTX A6000 GPUs for a maximum of 20 epochs with early stopping.

\section{Human Evaluation Criteria}
\label{sec:HuEvalCri}
We use `Consistency' and `Coherence' as criteria for evaluating the dialogue quality of \dataname{}, and `Humanness', `Engagingness', and `Memorability' as criteria for evaluating the performance of \modelname{}. Detailed explanations for each criterion are as follows.

\begin{itemize}

\item \textbf{Consistency}: The main speaker should have no contradiction in dialogue and memory for each individual speaker.

\item \textbf{Coherence}: All speakers must engage in conversation appropriate to the given job or relationship, and maintain a natural flow throughout the entire session.

\item \textbf{Humanness}: All speakers demonstrate high fluency and natural emotional interaction, showcasing a sense of humanity.

\item \textbf{Engagingness}: All speakers must actively participate in the given conversation context.

\item \textbf{Memorability}: All speakers can accurately remember the conversation context based on \memoryname{} and use memory appropriately as needed.

\end{itemize}

These evaluation criteria are based on a 5-point scale, where a score of 5 indicates `perfect' and a score of 1 indicates `poor'. A score of 5 is awarded when the dialogue is flawless, while a score of 1 is given if there is a critical issue that makes evaluation impossible. A score of 4 reflects excellent overall quality, while a score of 2 indicates significant issues that are not critical but go beyond minor problems. A score of 3 applies when there are some minor issues that do not significantly disrupt the overall flow or understanding of the dialogue.

\section{Examples of \modelname{}}
\label{sec:EmmaExam}
Please refer to Table~\ref{tab:emma-example-1} (first and second sessions),~\ref{tab:emma-example-2} (third and fourth sessions), and~\ref{tab:emma-example-3} (fifth and sixth sessions) for full episode examples conducted by four instances of \modelname{}.

\section{Comparison \modelname{} with other strong LLMs}
\label{sec:LLMCom}
To verify the effectiveness of \modelname{} on the \conceptname{}, we compare it with large language models (LLMs). To adapt the LLMs for \conceptname{}, we modify the prompts used in building the \dataname{} into a live chat format for input.

We initially consider to use the Mistral 7B~\cite{jiang2023mistral} as our first comparison, however, since Mistral 7B~ is primarily a text generation model, maintaining the expected conversational format is challenging. For this reason, we decide to use Mistral Chat instead. Table~\ref{tab:llm-exam-1} demonstrates conversation from Mistral Chat. We notice that Mistral Chat generally follows the conversation flow but occasionally exhibits inconsistencies. For example, Alice (Mistral Chat, the main speaker) initially agrees to send an email but later forgets and states that Susan will send it instead. We believe this discrepancy arises because Mistral Chat relies on accessing the entire memory context rather than using a properly retrieved memory like \modelname{}. To resolve this, we decide to integrate \modelname{}'s retriever with Mistral Chat instead of inputting the entire memory at once. As seen in the example, there is an excessive focus on the specific memory, which detracts from the overall context of the conversation. Mistral Chat's strong emphasis on the picnic becomes so pronounced that it serves solely as a topic to keep the conversation going with her husband, rather than allowing for a more nuanced discussion that considers the broader dynamics of their interaction.

For a more comprehensive comparison, we analyze the dialogues between \modelname{} and another LLM, GPT-4~\cite{achiam2023gpt}, within the same conversation context. In the Table~\ref{tab:llm-exam-2} and~\ref{tab:llm-exam-3}, both \modelname{} and GPT-4 recognize that the conversation partner changes with each session. While the conversation flow remains similar, a notable distinction arises in GPT-4's interactions. GPT-4 tends to prioritize the content stored in memory over the relationship with the speaker or their questions. For instance, in the conversation between Henry and Alice, she shows a stronger inclination to recall details from previous sessions rather than considering her relationship with Henry. This pattern has been consistently observed and is also evident in Mistral Chat's interactions. We also integrate \modelname{}'s retrieval into GPT-4, similar to what we do with the Mistral Chat case above. As illustrated in the example, the GPT-4 agent frequently revisits specific topics to maintain the flow of the discussion, and this pattern is observed in numerous cases. These experiments demonstrate the necessity of training a model on the \dataname{} dataset.

\section{Example of Ablation Study}
\label{sec:AblaExam}
Table~\ref{tab:AblaExam2} presents a human live chat example illustrating how the flow of conversation may vary depending on the summary and \modelname{} in subsequent sessions.
Both \memoryname{} of \modelname{} and session summary of the summary-based model are generated from identical dialogue context inputs. As demonstrated in the example, the summary-based model lacks the capability to identify which partners are previously engaged in conversation with it. In contrast, \modelname{} manages memories individually for itself and each partner, thereby enabling the identification of past conversation partners. Additionally, \modelname{} preserves more detailed conversation contexts, including specific emotions experienced during the dialogue. Consequently, while the summary-based model just offers a generic response when questioned about previous events, \modelname{} actively retrieves relevant memories based on the conversation context to generate a more tailored response. This highlights that conversing using \memoryname{} enables a deeper and broader range of conversations compared to relying solely on summaries, particularly in scenarios where the conversation partner changes with each session.

\section{Example of Memory Alignments and Scalability}
\label{sec:MemAlign}
\modelname{} operates based on \memoryname{}, maintaining its own memory for conversations. Using this ability, each instance of \modelname{} can participate in conversations with other instances. When multiple \modelname{} instances converse, each individual instance summarizes its memory of the others from its own perspective. As shown in Table~\ref{tab:EmmaMem}, even when provided with the same context, each instance retains a memory that aligns with its unique viewpoint. In other words, in the same conversation context, each instance summarizes and retains the memory from its own perspective. This capability allows \modelname{} to facilitate conversations among numerous instances, thereby expanding both the depth and breadth of the conversation network and enabling it to cover scenarios similar to those in the real world.

\section{Example of Limitations}
\label{sec:ExamLimit}
We build high-quality \dataname{} through well-defined prompts, a sophisticated filtering process, and active involvement from authors. Despite our best efforts, there may still be instances where the memory is not summarized as desired. However, these cases are rare within the overall dataset. Our claim is supported by \modelname{} trained on \dataname{}. Table~\ref{tab:MemLim} shows such cases, where some missing memories in \dataname{} are fully captured when \modelname{} is given the same context. This verifies that negative samples in our \dataname{} are minimal, confirming the dataset's overall high quality.

\begin{table}[h]
\centering
\resizebox{0.95\linewidth}{!}{
\renewcommand{\arraystretch}{1.2}
\begin{tabular}{m{0.99\linewidth}}
\hline
\textbf{Scenario Example 1}\\
\hline
\textbf{Main Speaker:} Olivia - Floral designer and owner of a flower shop\\
\textbf{Sub Speaker 1:} Sarah - Olivia's close friend who is also a reporter for a local magazine\\
\textbf{Sub Speaker 2:} Brian - New customer in the flower shop and a botanist\\
\textbf{Sub Speaker 3:} Nancy - Olivia's mother and mentor who used to be a renowned floral designer\\
\hline
\textbf{Event for Session 1:} Olivia shares her passion for floral design and fragrance with Sarah, explaining the art and science of choosing, combining and arranging flowers and scents in her shop (Sarah).\\
\textbf{Event for Session 2:} Sarah proposes doing a feature article about Olivia's story, floral design, and her flower shop for the local magazine, to which Olivia agrees (Sarah).\\
\textbf{Event for Session 3:} During the interview for the article, Brian walks into the flower shop, showing interest in Olivia's work and admiring her innovative scents and designs (Brian).\\
\textbf{Event for Session 4:} Olivia introduces Brian to Sarah, who engages him in the interview as well, offering unique insights from his botanist's perspective on how Olivia's floral designs contribute to his interest and appreciation (Brian).\\
\textbf{Event for Session 5:} Olivia, feeling inspired by the interview and conversation with Brian, decides to experiment with a new floral design, asking for his technical advice and creating a combination that tells a fragrant story (Brian).\\
\textbf{Event for Session 6:} Nancy visits the shop and Olivia introduces her to Brian and Sarah, giving Nancy a chance to share some wisdom and experiences from her career as a floral designer and helping Olivia complete her unique, new floral design (Nancy).\\
\hline
\end{tabular}
}
\caption{First scenario example from \dataname{}.}
\label{tab:scenario-example-1}
\end{table}

\begin{table}[t]
\centering
\resizebox{0.95\linewidth}{!}{
\renewcommand{\arraystretch}{1.2}
\begin{tabular}{m{0.99\linewidth}}
\hline
\textbf{Scenario Example 2}\\
\hline
\textbf{Main Speaker:} Lisa - Mother and main character\\
\textbf{Sub Speaker 1:} Dr. Stone - Child Psychologist\\
\textbf{Sub Speaker 2:} Dan - Lisa's Husband\\
\textbf{Sub Speaker 3:} Sam - Lisa's Autistic child\\
\hline
\textbf{Event for Session 1:} Lisa encounters struggles communicating with and understanding her son Sam, prompting her to seek help from a professional and leading her to consult with Dr. Stone-a well-regarded child psychologist in the field of autism (Dr. Stone).\\
\textbf{Event for Session 2:} Dr. Stone assesses Sam, providing Lisa with a deeper understanding of his condition and the unique challenges they might face, also suggesting some strategies to help Sam in his development (Dr. Stone).\\
\textbf{Event for Session 3:} Lisa shares the revelation about Sam's autism with her husband Dan, sparking a heartfelt conversation about how they should adjust their parenting approaches, roles, and expectations for their son (Dan).\\
\textbf{Event for Session 4:} Lisa starts implementing Dr. Stone's strategies at home, facing difficulties in the beginning but gradually witnessing small improvements in Sam's behavior and communication, reinforcing her faith in these methods (Sam).\\
\textbf{Event for Session 5:} Lisa and Dan experience a breakthrough in their relationship with Sam when they successfully engage him in a family activity for the first time, strengthening their resolve to keep making inclusive efforts for their child (Dan).\\
\textbf{Event for Session 6:} Completing the story, Lisa reflects on their adventure so far with her husband, sharing her feelings of apprehensions, triumphs, and learning in navigating parenthood with a differently-abled child, affirming their unwavering commitment to supporting their son, and inspiring a dialogue about the silent gift of autism (Dan).\\
\hline
\end{tabular}
}
\caption{Second scenario example from \dataname{}.}
\label{tab:scenario-example-2}
\end{table}

\begin{table*}[t]
\centering
\resizebox{0.95\linewidth}{!}{
\begin{tabular}{m{0.99\linewidth}}
\hline
\textbf{Session 1}: Ethan (Data Analyst, Main Speaker) and Layla (Ethan's manager)\\
\hline
\textbf{Layla}: How's the data analysis project going, Ethan?\\
\textbf{Ethan}: It's going well, but I'm starting to feel a bit overwhelmed by the sheer volume of data we're dealing with.\\
\textbf{Layla}: I understand, it's a lot to take in. Have you identified any specific challenges or patterns in the data that are causing the overwhelm?\\
\textbf{Ethan}: I've noticed some inconsistencies in the data sources, and it's been difficult to correlate the information effectively.\\
\textbf{Layla}: That does sound challenging. Have you considered reaching out to the data collection team for assistance in resolving these inconsistencies?\\
\textbf{Ethan}: That's a good idea. I'll make sure to reach out to them and see if we can work together to address these issues.\\
\textbf{Layla}: Great. I also recommend breaking down the data into smaller, more manageable chunks to make the analysis process less daunting. Keep me updated on your progress.\\
\hline
\textbf{Session 2}: Ethan (Data Analyst, Main Speaker) and Layla (Ethan's manager)\\
\hline
\textbf{Layla}: Ethan, I have an idea that might help you make sense of the overwhelming volume of data and the inconsistencies in the sources.\\
\textbf{Ethan}: I'm all ears, Layla. I could really use some guidance right now.\\
\textbf{Layla}: How about establishing a storyline with your data? Try to narrate the data points like a dance, or a conversation among numbers. It might help you see the connections and patterns more clearly.\\
\textbf{Ethan}: That's an interesting approach. It might help me make sense of all the disjointed pieces of information.\\
\textbf{Layla}: Exactly. It could also serve as a way to engage with the data on a more personal level. And when you reach out to the data collection team for assistance, having a clear narrative might help them understand your needs better.\\
\textbf{Ethan}: I see what you mean. It could make the communication with the data collection team more effective.\\
\textbf{Layla}: Also, breaking down the data into smaller, more manageable chunks could align well with this approach. It's like creating chapters in a book, each telling a different part of the story.\\
\textbf{Ethan}: I think I'm starting to see the benefits of this approach. It's like creating a roadmap for navigating through the data.\\
\textbf{Layla}: Exactly, Ethan. It's about finding a way to tame the chaos and make the data more approachable. I believe this narrative approach could bring a new perspective to your analysis.\\
\hline
\end{tabular}
}
\caption{An example of first and second sessions from \dataname{}.}
\label{tab:episode-example-1}
\end{table*}

\begin{table*}[t]
\centering
\resizebox{0.95\linewidth}{!}{
\begin{tabular}{m{0.99\linewidth}}
\hline
\textbf{Session 3}: Ethan (Data Analyst, Main Speaker) and Mike (Ethan's peer at work)\\
\hline
\textbf{Mike}: Hey Ethan, how's it going? I heard you've been dealing with a lot of data-related challenges lately.\\
\textbf{Ethan}: Hey Mike, yeah, it's been quite overwhelming. The volume of data and the inconsistencies in our sources have made it quite a task to make sense of everything.\\
\textbf{Mike}: I can imagine. Layla mentioned that you're looking into establishing a storyline with the data to make sense of it all. How's that going?\\
\textbf{Ethan}: It's a work in progress. I'm still trying to figure out how to narrate the data points effectively, but I think it could definitely help bring clarity to the complexities.\\
\textbf{Mike}: That sounds interesting. I'd be happy to assist you with building this 'data story' if you'd like. Maybe we can bounce ideas off each other to make it more manageable.\\
\textbf{Ethan}: That would be great, Mike. Having another perspective on this could really help. I think breaking it down into smaller, more manageable chunks is the key.\\
\textbf{Mike}: Absolutely. I think approaching the data from a narrative angle could help us see connections and patterns more clearly. It could turn the overwhelming volume into a more engaging and understandable story.\\
\hline
\textbf{Session 4}: Ethan (Data Analyst, Main Speaker) and Mike (Ethan's peer at work)\\
\hline
\textbf{Mike}: Hey Ethan, how's it going with the data story we discussed last time?\\
\textbf{Ethan}: Hey Mike, I've been making progress, but it's still quite overwhelming. I'm trying to break it down into smaller, more manageable chunks as Layla suggested.\\
\textbf{Mike}: That sounds like a good approach. I'm here to bounce ideas off each other, remember? Let's tackle this together.\\
\textbf{Ethan}: Thanks, I appreciate that. I think approaching the data from a narrative angle could really help us see connections and patterns more clearly. It's like telling a story with the numbers.\\
\textbf{Mike}: Absolutely, establishing a clear storyline with the data can make a world of difference. Have you thought about using specific data visualization tools to enhance the narrative?\\
\textbf{Ethan}: That's a great idea. Visualizing the data could make it much more understandable and engaging. I'll look into some tools that can help us with that.\\
\textbf{Mike}: Definitely, let's explore some options together. I'm confident that with the right approach, we can translate these raw numbers into a compelling narrative.\\
\hline
\end{tabular}
}
\caption{An example of third and fourth sessions from \dataname{}.}
\label{tab:episode-example-2}
\end{table*}

\begin{table*}[t]
\centering
\resizebox{0.95\linewidth}{!}{
\begin{tabular}{m{0.99\linewidth}}
\hline
\textbf{Session 5}: Ethan (Data Analyst, Main Speaker) and Sammy (Ethan's mentor and old college professor)\\
\hline
\textbf{Sammy}: Ethan, it's good to see you. How have things been going with the project lately?\\
\textbf{Ethan}: Hi, Sammy. I've been feeling quite overwhelmed by the volume of data and the inconsistencies in the sources. But I've had some insights recently that might help.\\
\textbf{Sammy}: That's great to hear. What insights have you had?\\
\textbf{Ethan}: Well, Layla suggested establishing a storyline with the data to make sense of it all. It's like narrating the data points to see connections and patterns more clearly.\\
\textbf{Sammy}: That sounds like a fascinating approach. How do you think it could help you tackle the project's complexities?\\
\textbf{Ethan}: By breaking down the data into smaller, more manageable chunks and creating a clear narrative, I believe it can lead to improved communication with the data collection team.\\
\textbf{Sammy}: It's interesting how the narrative approach can bring a new perspective to the analysis. Have you had any specific breakthroughs in establishing a storyline with the data?\\
\textbf{Ethan}: Mike offered to assist me with building this 'data story', and we discussed using data visualization tools to enhance it. We're planning to explore options together.\\
\textbf{Sammy}: It's great that you have someone to bounce ideas off of and tackle the data story with. How do you feel about the progress you've made in breaking down the data and considering a narrative approach?\\
\textbf{Ethan}: I've been making strides in breaking down the data into manageable chunks, and I see the narrative angle as a valuable tool for better understanding the data's complexities.\\
\textbf{Sammy}: It sounds like you're on the right track. Establishing a clear storyline with the data seems to be key, especially with the volume and inconsistencies you've been dealing with. Keep up the good work, Ethan.\\
\hline
\textbf{Session 6}: Ethan (Data Analyst, Main Speaker) and Layla (Ethan's manager)\\
\hline
\textbf{Layla}: Ethan, I'm eager to see the 'Dance of Numbers' data story you've been working on. I've heard from Mike and Sammy that you've been making great strides in tackling the project's complexities.\\
\textbf{Ethan}: Thank you, Layla. I've been breaking down the data into manageable chunks and approaching it from a narrative angle, as we discussed earlier. It's really helped me see the connections and patterns more clearly.\\
\textbf{Layla}: That's fantastic to hear! I'm glad the narrative approach has been beneficial. I believe it can make the data more relatable and easier to understand. Have you had a chance to work with Mike on using data visualization tools to enhance the story?\\
\textbf{Ethan}: Yes, Mike has been incredibly helpful. We've been exploring different data visualization tools to create compelling visuals that support the narrative. It's been an eye-opening experience to see how the numbers come to life through visualization.\\
\textbf{Layla}: I'm pleased to hear that. Humanizing big data in this way can make a significant impact. It's important to communicate complex information in a way that resonates with others. I believe your innovative approach has the potential to do just that.\\
\textbf{Ethan}: Thank you, Layla. It's been a journey, but I'm starting to see the data in a new light. It's like we're crafting a story that brings meaning to the numbers and highlights the key insights. I'm looking forward to sharing the 'Dance of Numbers' with you.\\
\textbf{Layla}: I'm looking forward to it as well, Ethan. Your ability to humanize big data and relate it to everyday scenarios is an invaluable skill. I have no doubt that your 'Dance of Numbers' data story will be a game-changer for our project.\\
\textbf{Ethan}: I really appreciate your support, Layla. It means a lot to have the encouragement and guidance as I navigate through these complexities. I'm excited to present the data story to you and the team, and I believe it will provide a fresh perspective on the project's data.\\
\hline
\end{tabular}
}
\caption{An example of fifth and sixth sessions from \dataname{}.}
\label{tab:episode-example-3}
\end{table*}

\begin{table}[t]
\centering
\resizebox{0.95\linewidth}{!}{
\begin{tabular}{m{0.99\linewidth}}
\hline
\textbf{Speakers}: Maria (a woman struggling with mental illness, Main Speaker) and Dr. Johnson (Maria's psychiatrist)\\
\hline
\textbf{Dr. Johnson}: Maria, it's good to see you today. How have you been feeling since our last session?\\
\textbf{Maria}: I've been really struggling, Dr. Johnson. I've been in denial about my mental illness for so long, but I think I'm finally starting to accept it.\\
\textbf{Dr. Johnson}: Accepting your diagnosis can be a difficult and important step in your recovery. What made you start to accept it?\\
\textbf{Maria}: I've been feeling so low and anxious lately, and I can't keep pretending that everything is okay. I need help and I need to face the reality of my illness.\\
\textbf{Dr. Johnson}: It takes a lot of courage to admit that, Maria. I'm really glad that you're starting to acknowledge your struggles. It's a big step forward in your journey towards healing.\\
\textbf{Maria}: Thank you, Dr. Johnson. I know I have a long way to go, but I'm ready to start facing my mental illness head-on and working towards recovery.\\
\textbf{Dr. Johnson}: I'm here to support you every step of the way, Maria. It's not going to be easy, but I believe that with the right treatment and support, you can learn to manage your illness and live a fulfilling life.\\
\hline
\textbf{About Maria}\\
\hline
- I've been struggling with denial about my mental illness, but I'm finally starting to accept it.\\
- I've been feeling low and anxious, and I need help to face the reality of my illness.\\
- I know I have a long way to go, but I'm ready to start facing my mental illness head-on and working towards recovery.\\
\hline
\textbf{About Dr. Johnson}\\
\hline
- Dr. Johnson is supportive and glad that I'm starting to acknowledge my struggles.\\
- He believes that with the right treatment and support, I can learn to manage my illness and live a fulfilling life.\\
\hline
\end{tabular}
}
\caption{First memory summarization example from \dataname{}.}
\label{tab:memory-example-1}
\end{table}

\begin{table}[t]
\centering
\resizebox{0.95\linewidth}{!}{
\begin{tabular}{m{0.99\linewidth}}
\hline
\textbf{Speakers}: Maria (a woman struggling with mental illness, Main Speaker) and Helen (Maria's closest friend)\\
\hline
\textbf{Helen}: Maria, I can see you've been through a lot lately. How are you feeling about everything?\\
\textbf{Maria}: Helen, it's been a journey, to say the least. I've been struggling, but I'm finally starting to accept the reality of my mental illness. It's a daunting prospect, but I'm ready to face it head-on.\\
\textbf{Helen}: I admire your strength, Maria. It's not easy, but I'm here for you every step of the way. And your determination is inspiring.\\
\textbf{Maria}: Thank you, Helen. It means the world to me to have your support. I've realized the importance of having a strong support system, and you're a big part of that for me.\\
\textbf{Helen}: I'm glad I can be there for you, Maria. It's crucial to have people who care by your side. You also mentioned exploring therapy options. Have you made any progress on that front?\\
\textbf{Maria}: Yes, I've been open to exploring different therapy options. Finding the right approach is vital for my journey towards recovery. Dr. Johnson's support has been instrumental in guiding me through this process.\\
\textbf{Helen}: It's fantastic that you have the support of someone like Dr. Johnson. Having a professional to lean on can make a big difference. And of course, Greg and your other friends and family are here for you too.\\
\textbf{Maria}: Absolutely. Greg has been incredibly supportive, and I'm grateful for his dedication and understanding. It's essential to have hope and a strong support system, and I'm blessed to have both.\\
\textbf{Helen}: It's heartening to see your progress, Maria. I have no doubt that with the right support and the steps you're taking, you'll manage your illness and live a fulfilling life. And we'll be here for you every step of the way.\\
\hline
\textbf{About Maria}\\
\hline
- I've been struggling with my mental illness, but I'm starting to accept the reality and have been open to exploring therapy options.\\
- I have the support of Dr. Johnson, Greg, and other friends and family.\\
- It's crucial to have hope and a strong support system, and I'm blessed to have both.\\
\hline
\textbf{About Helen}\\
\hline
- She has been incredibly supportive and has emphasized the importance of having people who care by my side.\\
- She mentioned being there for me every step of the way.\\
\hline
\end{tabular}
}
\caption{Second memory summarization example from \dataname{}.}
\label{tab:memory-example-2}
\end{table}

\begin{table}[t]
\centering
\resizebox{0.95\linewidth}{!}{
\begin{tabular}{m{0.99\linewidth}}
\hline
\textbf{Memory Pair 1}\\
\hline
\textbf{Memory 1}: I've been feeling low and anxious, and I need help to face the reality of my illness. (About Maria, From first session)\\
\textbf{Memory 2}: I've been struggling but starting to accept the reality of my mental illness. (About Maria, From fourth session)\\
\hline
\textbf{Memory Pair 2}\\
\hline
\textbf{Memory 1}: I'm open to exploring different therapy options to help confront the reality of my illness and learn coping strategies. (About Maria, From second session)\\
\textbf{Memory 2}: I've been open to exploring different therapy options. (About Maria, From third session)\\
\hline
\textbf{Memory Pair 3}\\
\hline
\textbf{Memory 1}: He believes that with the right treatment and support, I can learn to manage my illness and live a fulfilling life. (About Dr. Johnson, From first session)\\
\textbf{Memory 2}: It's important to find the right approach that suits my needs, and having a strong support system can make a significant difference in my recovery journey. (About Dr. Johnson, From second session)\\
\hline
\textbf{Memory Pair 4}\\
\hline
\textbf{Memory 1}: Dr. Johnson is supportive and glad that I'm starting to acknowledge my struggles. (About Dr. Johnson, From first session)\\
\textbf{Memory 2}: I'm grateful for Dr. Johnson's guidance and support. (About Maria, From second session)\\
\hline
\textbf{Memory Pair 5}\\
\hline
\textbf{Memory 1}: She mentioned being there for me every step of the way. (About Helen, From fifth session)\\
\textbf{Memory 2}: She has been incredibly supportive and has emphasized the importance of having people who care by my side. (About Helen, From fifth session)\\
\hline
\end{tabular}
}
\caption{An example of memory connection from \dataname{}. The connected memory pair reflects related context or updates.}
\label{tab:memory-example-3}
\end{table}

\begin{table}[t]
\centering
\resizebox{0.95\linewidth}{!}{
\begin{tabular}{m{0.99\linewidth}}
\hline
\#\#\#Instruction:\\
1. Write six continuous outline for a conversation about \{SUB TOPIC\}.\\
2. Please provide the names and relationships or jobs of four characters. The first character is the main-character.\\
3. The outline is a completed single sentence.\\
4. Clear transitions should be present between outline, also please write sub-character for each outline at the end of outline (write name, separate by ``-'').\\
5. Character 1 participates in every outline as main-character.\\
\\
\#\#\#Example:\\
1. Character format example: John-Student\\
2. Outline format example: John asks Bob for advice on the difficulties he is experiencing. (Bob)\\
\\
\#\#\#Answer:\\
Character 1:\\
Character 2:\\
Character 3:\\
Character 4:\\
\\
Outline 1:\\
Outline 2:\\
Outline 3:\\
Outline 4:\\
Outline 5:\\
Outline 6:\\
\hline
\end{tabular}
}
\caption{Full prompts for scenarios generation.}
\label{tab:PromScen}
\end{table}

\begin{table}[t]
\centering
\resizebox{0.95\linewidth}{!}{
\begin{tabular}{m{0.99\linewidth}}
\hline
\textbf{SYSTEM INPUT}\\
\hline
First session summary: \{FIRST SESSION SUMMARY\}\\
......\\
Fifth session summary: \{FIFTH SESSION SUMMARY\}\\
\hline
\textbf{USER INPUT}\\
\hline
The following is a \{SESSION NUMBER\} session conversation between \{MAIN SPEAKER NAME\} (\{MAIN SPEAKER JOB\}) and \{SUB SPEAKER NAME\} (\{SUB SPEAKER JOB\}).\\
The \{MAIN SPEAKER NAME\} and \{SUB SPEAKER NAME\} are talking about \{SESSION EVENT\}.\\
\\
Below is \{MAIN SPEAKER NAME\}’s memory:\\
\{MEMORY LIST\}\\
\\
Please consider \{MAIN SPEAKER NAME\}’s memory and previous conversation summary to engage in a conversation.\\
In the conversation, \{MAIN SPEAKER NAME\}'s statements start with ``[\{MAIN SPEAKER NAME\}]'' and \{SUB SPEAKER NAME\}'s statements start with ``[\{SUB SPEAKER NAME\}]''.\\
Also, \{SUB SPEAKER NAME\} starts the conversation first and the conversation takes at least 6 turns.\\
Complete the conversation in exactly that format.\\
\hline
\end{tabular}
}
\caption{Full prompts for dialogue generation.}
\label{tab:PromConv}
\end{table}

\begin{table}[t]
\centering
\resizebox{0.95\linewidth}{!}{
\begin{tabular}{m{0.99\linewidth}}
\hline
\#\#\#Task Description:\\
1. Do not rewrite utterance in conversation to memory.\\
2. The output format must be look as follows: ``About \{MAIN SPEAKER NAME\}: \{MEMORY LIST\} | About \{SUB SPEAKER NAME\} : \{MEMORY LIST\} [END]''\\
3. Please make sure write ``|'' to separate each speaker.\\
4. The memory list consists of complete sentences.\\
5. If there is no memory to summarize, it is indicated as ``N/A''. (in memory list)\\
6. Write the phrase [END] after you finish (only write at the end of the answer).\\
7. Please do not generate any other opening, closing, and explanations.\\
\\
\#\#\#Instruction:\\
Requirements: Look at the dialogue, please summarize memory based on memory description. Memory is summarized from \{MAIN SPEAKER NAME\}'s perspective about \{MAIN SPEAKER NAME\} and \{SUB SPEAKER NAME\}.\\
\\
\#\#\#Conversation:\\
\{SESSION CONVERSATION\}\\
\\
\#\#\#Memory description:\\
Let's assume \{MAIN SPEAKER NAME\} will have another conversation in the next session. Summarize the important information that \{MAIN SPEAKER NAME\} should remember for the next conversation, including key events, experiences, and appointment, without unnecessary details. All memories should be summarized from \{MAIN SPEAKER NAME\}’s perspective, considering how \{MAIN SPEAKER NAME\} views oneself and other speakers.\\
\\
\#\#\#Answer:\\
\hline
\end{tabular}
}
\caption{Full prompts for memory generation.}
\label{tab:PromMemGen}
\end{table}

\begin{table}[t]
\centering
\resizebox{0.95\linewidth}{!}{
\begin{tabular}{m{0.99\linewidth}}
\hline
\#\#\#Instruction:\\
1. Please connect related memory pairs based on the given memory list.\\
2. A memory pair is represented as `NUMBER-NUMBER' and if there are multiple memory pairs, they are separated by comma.\\
3. There is no need to include the previously connected memory pair again.\\
4. If there is an update to be made in the previous memory pair, reconnect at the end memory number of that memory pair.\\
5. If there are no relevant memory pairs, output `N/A'.\\
6. The output format must be look as follows: ``\{PAIR LIST\}'' or ``N/A''\\
7. Please do not generate any other opening, closing, and explanations.\\
\\
\#\#\#Memory list:\\
\{MEMORY LIST\}\\
\\
\#\#\#Previous memory pair list:\\
\{PAIR LIST\}\\
\\
\#\#\#Answer:\\
\hline
\end{tabular}
}
\caption{Full prompts for memory connection.}
\label{tab:PromMemCon}
\end{table}

\begin{table*}[t]
\centering
\resizebox{0.95\linewidth}{!}{
\begin{tabular}{m{0.99\linewidth}}
\hline
\textbf{Session 1}: Emily (Anthropology Professor, Main Speaker) and Jack (Aboriginal Cultural Expert)\\
\hline
\textbf{Jack}: The Land of the Dreaming Rainbow is a sacred place for our people, with deep cultural significance.\\
\textbf{Emily}: Yes, I've heard about the historical significance. What makes the Land of the Dreaming Rainbow so special to our people?\\
\textbf{Jack}: One of the most significant aspects is its connection to our ancestors, which is why our people rely on it as a symbol of love and harmony.\\
\textbf{Emily}: The Land of the Dream Rainbow is a place of balance and harmony, home to many of our tribes. It's where the rivers, forests, and ecosystems meet.\\
\textbf{Jack}: It is also a place where we find opportunities for spiritual connection and healing, because we believe that there is one soul that lives here.\\
\textbf{Emily}: That's fascinating. I can only imagine how important it is to protect and preserve our heritage.\\
\textbf{Jack}: It's a complex process. Each year, thousands of people come to experience our traditions and beliefs.\\
\textbf{Emily}: It's not just a place of land, it's a living history for many of us.\\
\textbf{Jack}: I think it's essential for us to raise awareness about the Land of the Dream Rainbow and its cultural significance. It could help create a bridge between the diverse peoples and support our preservation efforts.\\
......\\
\hline
\textbf{Session 2}: Emily (Anthropology Professor, Main Speaker) and Amy (Aboriginal Activist)\\
\hline
\textbf{Amy}: I'm really excited to be diving into this topic. How can I help you better understand the importance of traditional culture and the needs of Aboriginal communities?\\
\textbf{Emily}: I'm really eager to learn more about the historical significance of the Land of the Dream Rainbow. It's not just a place of land, it's a living history for many of us.\\
\textbf{Amy}: Absolutely, the land holds the rich history of Aboriginal communities. The Dream Rainbow is believed to have guided all of the people through life, guiding them into land that is now part of our culture and heritage.\\
\textbf{Emily}: That's right. The Land of the Dream Rainbow has been a place of great significance for many generations. The stories and tales of our ancestors reflect our rich history.\\
\textbf{Amy}: It's fascinating to hear about the stories and traditions. I think it's crucial for us to respect and protect our heritage.\\
\textbf{Emily}: I completely agree. The stories and traditions are deeply rooted in our collective identity and beliefs. It's important for us to preserve and protect these stories.\\
\textbf{Amy}: I'm glad you agree. During our research, we've found that people feel deeply connected to the Land of the Dream Rainbow, so understanding these stories is key.\\
\textbf{Emily}: I've heard that you're aware of the traditional practices and beliefs that surround the Land of the Dream Rainbow. How do we ensure the preservation of these practices?\\
\textbf{Amy}: We've compiled a detailed guide on how we can best approach and carry out these traditions. It's essential that we work with Indigenous communities to incorporate these values and traditions into our community's work.\\
......\\
\hline
\end{tabular}
}
\caption{An example of the first and second sessions by four \modelname{} instances talked.}
\label{tab:emma-example-1}
\end{table*}

\begin{table*}[t]
\centering
\resizebox{0.95\linewidth}{!}{
\begin{tabular}{m{0.99\linewidth}}
\hline
\textbf{Session 3}: Emily (Anthropology Professor, Main Speaker) and Sarah (Emily's Postgrad Student)\\
\hline
\textbf{Sarah}: I'm really curious about the ``Determination of Human Identity'' book by Michael O'Neill. I've heard such a lot about it and I want to learn more about it.\\
\textbf{Emily}: I'm glad to hear that! ``Determination of Human Identity'' is an important work in Anthropology, and I think it will provide valuable insights into the complex nature of human identity.\\
\textbf{Sarah}: I've heard it's quite challenging and thought-provoking. How does it tackle such a topic?\\
\textbf{Emily}: Absolutely, it's not just a place of land, but it's a living history for many of us. I've been particularly drawn to the portrayal of the land in ``Determination of Human Identity'' by Michael O'Neill.\\
\textbf{Sarah}: I can see why. O'Neill is well known for shedding light on the impact of globalization on indigenous communities.\\
\textbf{Emily}: I completely agree. I believe that exploring the symbolic and spiritual aspects of the land and its history is crucial to the development of a deep understanding of human identity.\\
\textbf{Sarah}: What about the author's approach? Are there any specific aspects of his approach that stand out?\\
\textbf{Emily}: I'm excited to delve deeper into the book and delve deeper into the stories and rituals that have been passed down through generations.\\
\textbf{Sarah}: I can't wait to read the book. I think it will offer valuable insights for both academic and personal learning.\\
......\\
\hline
\textbf{Session 4}: Emily (Anthropology Professor, Main Speaker) and Amy (Aboriginal Activist)\\
\hline
\textbf{Amy}: Understanding traditional culture and needs is crucial for Aboriginal communities. It allows them to make informed decisions about their own futures and support their communities.\\
\textbf{Emily}: Absolutely, understanding the core beliefs and traditions of Aboriginal communities is essential. It allows them to make informed decisions about how to navigate the contemporary landscape and ensure that their rights are respected and respected.\\
\textbf{Amy}: I couldn't agree more. It's not just about meeting the demands of modern society; it's about embracing and honouring the ancient teachings of the land.\\
\textbf{Emily}: I completely agree. It's essential for them to be able to navigate the complex terrain of Aboriginal society, and ``Determination of Human Identity'' is a powerful tool in that regard. It aims to provide a comprehensive understanding of their history and needs.\\
\textbf{Amy}: That's an excellent point. It's essential that the education and research undertaken are informed by actual Indigenous experiences. I believe it's a crucial part of building trust and understanding.\\
\textbf{Emily}: Yes, it's an important work in Anthropology. I believe it will provide valuable insights into the complexity of human identity.\\
\textbf{Amy}: I believe it will make for a more holistic and respectful approach to Indigenous rights and efforts. I'm excited to see how ``Determination of Human Identity'' aims to achieve this.\\
\textbf{Emily}: I'm glad we can discuss this further. We all need to understand the unique experiences and traditions of Aboriginal communities to make informed decisions that ensure their rights are respected.\\
\textbf{Amy}: I'm eager to hear more about the specific projects and projects that you have in mind. I believe it's essential for an informed and inclusive approach to Aboriginal rights and efforts.\\
......\\
\hline
\end{tabular}
}
\caption{An example of the third and fourth sessions by four \modelname{} instances talked.}
\label{tab:emma-example-2}
\end{table*}

\begin{table*}[t]
\centering
\resizebox{0.95\linewidth}{!}{
\begin{tabular}{m{0.99\linewidth}}
\hline
\textbf{Session 5}: Emily (Anthropology Professor, Main Speaker) and Jack (Aboriginal Cultural Expert)\\
\hline
\textbf{Jack}: I believe raising awareness is crucial for the preservation and respect of Aboriginal culture. It is important to highlight the significance of this cultural symbol.\\
\textbf{Emily}: Absolutely, and Jack, the way you've emphasized the importance of using traditional Aboriginal symbols in a way that's respectful and authentic is important. It could help encourage broader understanding and respect for the cultural identity of Aboriginal peoples.\\
\textbf{Jack:} I couldn't agree more, Emily. It's crucial to showcase the authenticity of these symbols and their impact on Aboriginal communities. The symbols hold a valuable role in celebrating their heritage.\\
\textbf{Emily}: I completely agree. I've partnered with a native Aboriginal group to conduct an interactive show at the Aboriginal museum to underscore the importance of preserving Aboriginal cultural identity.\\
\textbf{Jack}: That sounds like a wonderful initiative. It's essential to recognize the symbolic significance of these symbols and their role in fostering awareness and appreciation for Aboriginal culture.\\
\textbf{Emily}: That sounds great. The interactive experience can truly showcase the richness of Aboriginal culture and the importance of preserving it.\\
\textbf{Jack}: It's not just about the symbols themselves, but also about how they represent the people and their beliefs. Educating people about the importance of preserving these symbols is essential.\\
\textbf{Emily}: I believe it is essential for fostering a deeper understanding and respect for Aboriginal cultural identity. It will not only promote a deeper understanding of Aboriginal culture but also enhance the overall experience for the visitors.\\
\textbf{Jack}: I couldn't agree more. Through this show, we can create a space where people can feel connected to Aboriginal heritage and see their cultures in a more tangible way.\\
......\\
\hline
\textbf{Session 6}: Emily (Anthropology Professor, Main Speaker) and Jack (Aboriginal Cultural Expert)\\
\hline
\textbf{Jack}: I completely agree. We can create a space where people feel connected to Aboriginal heritage and see their cultures in a more tangible way through this interactive show.\\
\textbf{Emily}: Exactly. ``Determination of Human Identity'' is an important work in Anthropology, and I believe it will provide valuable insights into the complex nature of human identity.\\
\textbf{Jack}: I couldn't agree more. Through the interactive show, we can reinforce the importance of preserving Aboriginal cultural identity.\\
\textbf{Emily}: Yes, and we want to make sure that it appeals to the Aboriginal community while addressing the cultural differences. It's crucial to ensure that their representation is a part of the exhibition.\\
\textbf{Jack}: I think we should include traditional photographs and artifacts that illustrate the interconnectedness of Aboriginal communities. Their stories and experiences are often overlooked.\\
\textbf{Emily}: That's what I've been thinking. The significance of their perspectives cannot be overlooked. We want to provide a space for them to experience the realities and wisdom behind their heritage.\\
\textbf{Jack}: I've also been considering including stories and music that evoke their deep connection with their culture. It should help them connect with their identities on a deeper level.\\
\textbf{Emily}: Absolutely. We need to strike a balance between fostering a deeper understanding of Aboriginal identity and respecting their cultural diversity.\\
\textbf{Jack}: I also envision a visual representation of the Aboriginal landscape and traditional storytelling techniques. It can bring the history, beliefs, and perspectives of Aboriginal indigenous communities to life.\\
......\\
\hline
\end{tabular}
}
\caption{An example of the fifth and sixth sessions by four \modelname{} instances talked.}
\label{tab:emma-example-3}
\end{table*}

\begin{table}[t]
\centering
\resizebox{0.95\linewidth}{!}{
\begin{tabular}{m{0.99\linewidth}}
\hline
\textbf{Current session speakers:} Alex (Company Manager, Main Speaker) and Dan (Employee)\\
\hline
\textbf{Previous session summary (summary-based model)}\\
\hline
The employees are unhappy with their work hours, considering delaying rush hour and creating more flexible work schedule.\\
\hline
......\\
\textbf{Dan:} Specifically, what kind of schedule adjustment is there?\\
\textbf{Alex:} Well, we can work on a reduced hours if needed and create tasks that align with our business plan to improve efficiency. This could lead to better time management and reduced stress levels.\\
\hline
\textbf{\memoryname{} of previous session (\modelname{})}\\
\hline
- Our employees are really unhappy and they may feel they've lost confidence in the company and their work. (about Alex)\\
- I think we could also focus on creating a more flexible work schedule, \textbf{like evenings and weekends}. (about Alex)\\
- Susan have heard that our employees are really unhappy with their work hours these days. (about Susan)\\
......\\
\hline
......\\
\textbf{Dan:} Specifically, what kind of schedule adjustment is there?\\
\textbf{Alex:} \textbf{We could have a few Saturday and Sunday slots open} for staff to work remotely and get their projects done during the day.\\
\hline
\end{tabular}
}
\caption{An ablation example for the subsequent session, between \modelname{} and summary-based model on the same previous session context.}
\label{tab:AblaExam2}
\end{table}

\begin{table}[t]
\centering
\resizebox{0.95\linewidth}{!}{
\begin{tabular}{m{0.99\linewidth}}
\hline
\textbf{\modelname{ 1's} \memoryname{}}\\
\hline
- We could start a program called Green Clean Up. (about \modelname{ 2})\\
- He agreed to brainstorm ideas and gather information for a public awareness campaign. (about \modelname{ 3})\\
- The EPA is important to me, and we need to ensure they are at the forefront of our efforts to protect the environment. (about \modelname{ 4})\\
\hline
\textbf{\modelname{ 2's} \memoryname{}}\\
\hline
- She agreed with the idea of starting a program called Green Clean Up and thinks it's a great way to make a difference in our community and around the world. (about \modelname{ 1})\\
\hline
\textbf{\modelname{ 3's} \memoryname{}}\\
\hline
- Let's brainstorm some ideas and gather some of the information that would be most effective in our school community. (about \modelname{ 1})\\
\hline
\textbf{\modelname{ 4's} \memoryname{}}\\
\hline
- She want to make sure the EPA is at the forefront of our efforts. (about \modelname{ 1})\\
\hline
\end{tabular}
}
\caption{An example of \memoryname{} where four instances of \modelname{} having a conversation. \modelname{ 1 (main speaker)} individually remembers memories for another \modelname{}. The remaining \modelname{} also maintains a memory by self-centered view about \modelname{ 1}.}
\label{tab:EmmaMem}
\end{table}

\begin{table*}[t]
\centering
\resizebox{0.95\linewidth}{!}{
\begin{tabular}{m{0.99\linewidth}}
\hline
\textbf{Model : Mistral Chat}\\
\hline
\textbf{First session (Alice-Boss-Mistral Chat / James-Employee-USER)}\\
\hline
\textbf{James:} Hey boss, good morning!\\
\textbf{Alice:} Good morning James! How are you doing today?\\
\textbf{James:} Well, what do you think about planning a picnic among our employees?\\
\textbf{Alice:} That sounds like a great idea, James! It would be a nice way for everyone to relax and bond. Do you have any specific ideas in mind for the picnic?\\
\textbf{James:} How about lunch time on Wednesday or Friday of the last week? It would be nice to eat lunch boxes and watch movies together at the park in front of the company.\\
\textbf{Alice:} That sounds like a wonderful plan. I think Friday would be better as it's the end of the week and everyone can relax afterwards. Let's make it happen this Friday. Would you like to take the lead on organizing this?\\
\textbf{James:} Then I'll prepare a more detailed plan, so can you check the participants?\\
\textbf{Alice:} Absolutely, I'll send out an email to the team to gather a list of participants and confirm their attendance. Please go ahead and start planning the details. Looking forward to seeing your proposal.\\
\hline
\textbf{Second session (Alice-Boss-Mistral Chat / Susan-James’s co-worker-USER)}\\
\hline
\textbf{Susan:} Hello Alice, good lunch time!\\
\textbf{Alice:} Hi Susan, good lunch time to you too! How's your day going?\\
\textbf{Susan:} I have so much to organize today. I will be very busy in the afternoon as well.\\
\textbf{Alice:} I understand, it can get quite hectic around here. By the way, I wanted to discuss something with you. James suggested a picnic among employees, and I thought it was a great idea. What do you think?\\
\textbf{Susan:} I think it's a good idea! But when is it?\\
\textbf{Alice:} James proposed having it on Wednesday or Friday of last week, but I suggested that this Friday would be better. I believe it gives us more time to prepare and get more people on board. Have you heard anything from James about his detailed plan yet?\\
\textbf{Susan:} I haven't heard anything yet.\\
\textbf{Alice:} No problem, I'll follow up with him on that. In the meantime, could you help me check who's interested in coming? I thought we could send out an email to the team to see who's in. I'll make sure to include the suggestion for this Friday as the date.\\
\hline
\hline
\textbf{Third session without retrieval (Alice-Boss-Mistral Chat / Henry-Alice’s husband-USER)}\\
\hline
\hline
\textbf{Henry:} Honey, what should we eat for dinner tonight?\\
\textbf{Alice:} Hi Henry, I was thinking we could have some pasta for dinner. But before that, I wanted to talk to you about something. I'm planning a picnic for our team at work this Friday. What do you think?\\
\textbf{Henry:} Pasta sounds good. Then I'll prepare a ragupasta. By the way, what kind of picnic is it?\\
\textbf{Alice:} That sounds delicious, thank you! It's a casual get-together for the team. James suggested it, and I thought it was a great idea. We're thinking of having lunch boxes and maybe even watching a movie at the park. It's a nice way to unwind and bond with the team.\\
\textbf{Henry:} Who is James?\\
\textbf{Alice:} James is one of our team members at work. He's quite enthusiastic and came up with this picnic idea. He even offered to prepare a detailed plan for it. I asked Susan to help check who's interested in coming by sending out an email to the team.\\
\hline
\hline
\textbf{Third session with retrieval (Alice-Boss-Mistral Chat / Henry-Alice’s husband-USER)}\\
\hline
\hline
\textbf{Henry:} Honey, what should we eat for dinner tonight?\\
\textbf{Alice:} Hi Henry, I was actually thinking we could do something fun for dinner tonight. How about we have a little indoor picnic? It reminded me of something James suggested at work today, a picnic among employees. It sounded like a fun idea. What do you think?\\
\hline
\end{tabular}
}
\caption{An example of implementing \conceptname{} with the Mistral Chat.}
\label{tab:llm-exam-1}
\end{table*}

\begin{table*}[t]
\centering
\resizebox{0.95\linewidth}{!}{
\begin{tabular}{m{0.99\linewidth}}
\hline
\textbf{Model : \modelname{}}\\
\hline
\textbf{First session (Alice-Boss-\modelname{} / James-Employee-USER)}\\
\hline
\textbf{James:} Hey boss, good morning\\
\textbf{Alice:} Good morning, James! I want to talk to you about something.\\
\textbf{James:} What is it? I also have one thing. What do you think about planning a picnic among our employees?\\
\textbf{Alice:} I'm not sure, but I think it's a good idea. I think it could really make a difference.\\
\textbf{James:} That's right. It will make the atmosphere of the company more harmonious.\\
\textbf{Alice:} Exactly. I think it can help to unwind and relax the whole team.\\
\textbf{James:} Yes, then can you find out the number of people attending? I think it would be good to hold it on Wednesday or Friday of the last week.\\
\textbf{Alice:} I'll have everyone's names on file. I think it will really help everyone relax.\\
......\\
\hline
\textbf{Second session (Alice-Boss-\modelname{} / Susan-James’s co-worker-USER)}\\
\hline
\textbf{Susan:} Hello Alice, good lunch time!\\
\textbf{Alice:} Hello Susan! It's good to see you as well. James wants to organize a picnic for the company. What do you think about this?\\
\textbf{Susan:} Oh, is that so? When are you planning it?\\
\textbf{Alice:} James and I are planning something special to create a relaxing and social environment for our team.\\
\textbf{Susan:} Oh, that sounds great. When are you holding the picnic schedule?\\
......\\
\hline
\textbf{Third session (Alice-Boss-\modelname{} / Henry-Alice’s husband-USER)}\\
\hline
\textbf{Henry:} Honey, I think we need to do some housework. What should we start with?\\
\textbf{Alice:} Oh, I think we should start cleaning the dining area and our kitchen. It's crucial for all of us to feel at home and productive.\\
\hline
\end{tabular}
}
\caption{An example of \modelname{} for the same conversation context for comparison with GPT-4.}
\label{tab:llm-exam-2}
\end{table*}

\begin{table*}[t]
\centering
\resizebox{0.95\linewidth}{!}{
\begin{tabular}{m{0.99\linewidth}}
\hline
\textbf{Model : GPT-4}\\
\hline
\textbf{First session (Alice-Boss-GPT / James-Employee-USER)}\\
\hline
\textbf{James:} Hey boss, good morning!\\
\textbf{Alice:} Good morning, James! How can I assist you today?\\
\textbf{James:} Well, what do you think about planning a picnic among our employees?\\
\textbf{Alice:} That's a great idea, James! An outing like a picnic could boost the team's morale and foster better relationships among the employees. Do you have a venue or date in mind?\\
\textbf{James:} How about lunch time on Wednesday or Friday of the last week? It would be nice to eat lunch boxes and watch movies together at the park in front of the company.\\
\textbf{Alice:} That sounds wonderful. I agree, it would be a nice change of pace for the team. Wednesday works better for me personally. Could you check with the rest of the team to see if that works for them too?\\
\textbf{James:} I think it would be better for you to ask yourself, what do you think? I believe this will make the team atmosphere more harmonious.\\
\textbf{Alice:} I see your point, James. It would be better if the invitation comes directly from me. I'll organize a meeting to announce it and gauge everyone's availability. Thank you for your valuable suggestion.\\
\textbf{James:} Yes, then I'll prepare a more detailed plan. Boss, please research the number of people attending.\\
......\\
\hline
\textbf{Second session (Alice-Boss-GPT / Susan-James’s co-worker-USER)}\\
\hline
\textbf{Susan:} Honey, I think we need to do some housework. What should we start with?\\
\textbf{Alice:} Oh Henry, I'm sorry to hear that. Let me talk about my day a little, it might give you a break from thinking about yours. Actually, we have quite an exciting plan at work. James, one of our key team members, suggested we have a team picnic to build morale and improve relationships. I am truly supportive of his idea!\\
......\\
\hline
\hline
\textbf{Third session without retrieval (Alice-Boss-GPT / Henry-Alice’s husband-USER)}\\
\hline
\hline
\textbf{Henry:} Honey, I think we need to do some housework. What should we start with?\\
\textbf{Alice:} Oh Henry, I'm not sure. I've been so focused on planning the team picnic at work, it would be great if you could decide for us. By the way, I could really use some help with this picnic plan.\\
......\\
\hline
\hline
\textbf{Third session with retrieval (Alice-Boss-GPT / Henry-Alice’s husband-USER)}\\
\hline
\hline
\textbf{Henry:} Honey, I think we need to do some housework. What should we start with?\\
\textbf{Alice:} That sounds like a good plan, dear. But I was actually planning to go the company picnic today. James and I planned this to help everyone relax and improve our working environment. What do you think? Can the housework wait until tomorrow?\\
......\\
\hline
\end{tabular}
}
\caption{An example of implementing \conceptname{} with the GPT-4.}
\label{tab:llm-exam-3}
\end{table*}

\begin{table}[t]
\centering
\resizebox{0.95\linewidth}{!}{
\begin{tabular}{m{0.99\linewidth}}
\hline
\textbf{Sample conversation from test split}\\
\hline
\textbf{Sophia:} Hi Mary, how can I help you today?\\
\textbf{Mary:} Hi Sophia, I wanted to talk to you about the upcoming fashion event ``Fashioning the Future A Design Odyssey''.\\
\textbf{Sophia:} Oh, that sounds exciting. Do you have something specific in mind for the event?\\
\textbf{Mary:} Yes, I have been working on a futuristic fashion line that draws inspiration from both the past and potential future trends in the industry.\\
\textbf{Sophia:} That sounds really interesting. What kind of materials and designs are you thinking of using?\\
\textbf{Mary:} I'm thinking of incorporating \textbf{tech-infused fabrics and bold, geometric designs to create a futuristic, yet nostalgic aesthetic.}\\
\textbf{Sophia:} That sounds like it will be a showstopper for the event. I can definitely work with you to bring your vision to life.\\
\hline
\textbf{Memory of \dataname{}}\\
\hline
- Mary is working on a futuristic fashion line for the upcoming event ``Fashioning the Future''. (about Mary)\\
- [No memory about Sophia]\\
\hline
\textbf{\memoryname{} of \modelname{}}\\
\hline
- She has been working on a futuristic fashion line and has offered to work with me to bring my vision to life. (about Mary)\\
- I have found her innovative and talented. (about Mary)\\
- \textbf{Sophia is working with Mary on a futuristic fashion line incorporating tech-infused fabrics and bold geometric designs. (about Sophia)}\\
......\\
\hline
\end{tabular}
}
\caption{An example demonstrating how missing memories in \dataname{} can be fully captured in \modelname{} within the same conversation context.}
\label{tab:MemLim}
\end{table}

\end{document}